%% file: main.tex
\documentclass[10pt,twocolumn,letterpaper]{article}
\pdfoutput=1

\usepackage[pagenumbers]{cvpr} 


\definecolor{cvprblue}{rgb}{0.21,0.49,0.74}
\usepackage[pagebackref,breaklinks,colorlinks,allcolors=cvprblue]{hyperref}

\usepackage{booktabs}
\usepackage{multirow}
\usepackage{amsmath}
\usepackage{graphicx}
\usepackage{bbding}
\usepackage{algorithm}
\usepackage{algorithmic}

\usepackage[marginal]{footmisc}
\usepackage{color}

\def\paperID{3251} 
\def\confName{CVPR}
\def\confYear{2025}

\title{Satellite Observations Guided Diffusion Model for Accurate Meteorological States at Arbitrary Resolution}

\makeatletter
\renewcommand\@makefntext[1]{\noindent\makebox[1em][r]{\textsuperscript{$\dagger$}}#1}
\makeatother

\author{Siwei Tu$^{1,2}$, Ben Fei$^{2,3\dagger}$, Weidong Yang{$^{1\dagger}$}, Fenghua Ling$^{2}$, Hao Chen$^{2}$, Zili Liu$^{4}$, Kun Chen$^{1,2}$ \\
Hang Fan$^{2,5,6}$, Wanli Ouyang$^{2,3}$, Lei Bai$^{2}$ \\
	$^1$Fudan University $^2$Shanghai Artificial Intelligence Laboratory \\
    $^3$The Chinese University of Hong Kong $^4$Beihang University \\
    $^5$Columbia University $^6$Nanjing University of Information Science and Technology \\
}

\begin{document}
\maketitle
\input{sec/0_abstract} 

\footnotetext{Corresponding Authors: Ben Fei (\href{benfei@cuhk.edu.hk)}{benfei@cuhk.edu.hk)} and Weidong Yang (\href{wdyang@fudan.edu.cn}{wdyang@fudan.edu.cn})}

\input{sec/1_introduction}
\input{sec/2_related}
\input{sec/3_method}
\input{sec/4_experiment}

\input{sec/5_conclusion}


{
    \small
    \bibliographystyle{ieeenat_fullname}
    \bibliography{main}
}

\input{sec/X_suppl}

\end{document}

%% file: sec/0_abstract.tex
\begin{abstract}
Accurate acquisition of surface meteorological conditions at arbitrary locations holds significant importance for weather forecasting and climate simulation. 
Due to the fact that meteorological states derived from satellite observations are often provided in the form of low-resolution grid fields, the direct application of spatial interpolation to obtain meteorological states for specific locations often results in significant discrepancies when compared to actual observations. 
Existing downscaling methods for acquiring meteorological state information at higher resolutions commonly overlook the correlation with satellite observations. 
To bridge the gap, we propose \textbf{S}atellite-observations \textbf{G}uided \textbf{D}iffusion Model (\textbf{SGD}), a conditional diffusion model pre-trained on ERA5 reanalysis data with satellite observations (GridSat) as conditions, which is employed for sampling downscaled meteorological states through a zero-shot guided sampling strategy and patch-based methods.
During the training process, we propose to fuse the information from GridSat satellite observations into ERA5 maps via the attention mechanism, enabling SGD to generate atmospheric states that align more accurately with actual conditions. 
In the sampling, we employed optimizable convolutional kernels to simulate the upscale process, thereby generating high-resolution ERA5 maps using low-resolution ERA5 maps as well as observations from weather stations as guidance.  
Moreover, our devised patch-based method promotes SGD to generate meteorological states at arbitrary resolutions.
Experiments demonstrate SGD fulfills accurate meteorological states downscaling to 6.25km. 
\end{abstract}

%% file: sec/1_introduction.tex
\section{Introduction}
\label{sec:intro}

Precision in acquiring meteorological variable states at a small scale is pivotal to weather forecasting, which endeavors to predict forthcoming meteorological conditions by correlating current weather phenomena with subsequent atmospheric states, further aiding citizens, businesses, and nations in making informed decisions regarding future societal activities~\cite{mukkavilli2023ai,chen2023foundation,li2024deepphysinet}. 
Therefore, enhancing the accuracy of weather forecasts holds considerable significance for enterprise production and daily life, which necessitates extracting more accurate meteorological data on a finer scale~\cite{liu2024deriving,liu2024mambads}.

\begin{figure}[t]
    \centering
\includegraphics[width=\linewidth]{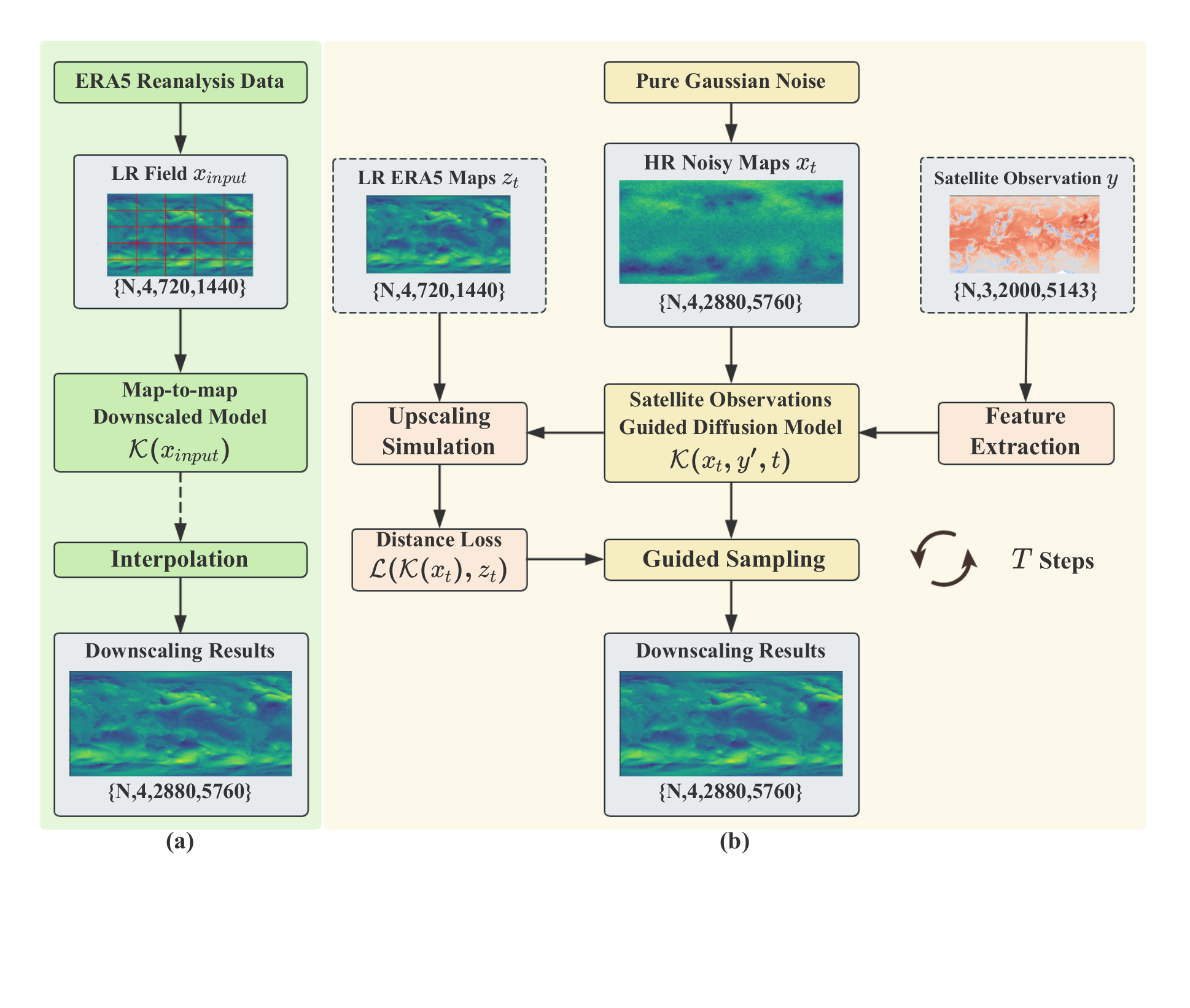}
    \vspace{-1.6cm}
    \caption{
    The difference between \textbf{(a)} the previous super-resolution (SR)-based, interpolation-based downscaling methods and \textbf{(b)} the proposed  SGD. 
    The SR-based methods attempt to model the downscaling process directly from low-resolution (LR) maps, while the interpolation-based methods solely rely on interpolation.
    However, these approaches introduce systematic biases and the loss of detail when dealing with maps at a small scale of $6.25km$. 
    In contrast, SGD endeavors to commence with high-resolution (HR) maps, employing satellite observations to conditionally sample via a diffusion model. 
    Simultaneously, it simulates and constructs the inverse process of downscaling by utilizing the original LR ERA5 maps and observations from weather stations, thereby guiding the sampling results to ensure the fidelity of detailed information. }
    \label{fig:teaser}
\vspace{-0.5cm}
\end{figure}

Downscaling is an effective method for capturing details meteorological data at a finer scale, which aims to transform low-resolution maps from meteorological data such as ERA5 reanalysis dataset into corresponding high-resolution maps at a small scale to obtain more accurate and detailed meteorological data~\cite{fang2013spatial, de2018estimating, aich2024conditional}. 
As shown in \cref{fig:teaser} \textbf{(a)}, in the task of downscaling meteorological data, spatial interpolation-based methods such as linear and bilinear interpolations are common and feasible approaches. 
These methods do not use learnable parameters to model the downscaling process but instead obtain meteorological states at the detail level of small-scale maps through interpolation from grid meteorological field data. 
Consequently, achieving a highly satisfactory level of precision in the downscaling process is challenging when dealing with complex and high-resolution grid information. 
In recent years, artificial intelligence technologies have demonstrated remarkable performance in this task~\cite{sun2024deep,wang2021method}. 
For instance, SwinRDM~\cite{chen2023swinrdm} utilizes diffusion models to recover high spatial resolution and finer-scale atmospheric details. 
\cite{pozo2021dynamically} utilized a high-resolution regional ocean circulation model to dynamically downscale Earth System Models (ESMs) and produce climate projections for the California Current System. 

Existing methods rely only on the original meteorological field data at relatively coarse resolutions and directly construct and model the downscaling process~\cite{hess2024fast, zhu2024downscaling}. 
However, for the downscaling task of ERA5 maps, a coupling relationship exists between ERA5 reanalysis data and satellite observations~\cite{vaughan2024aardvark,vandal2024global}. 
Satellite observations, such as brightness temperature, and humidity, are the major factors influencing the atmospheric states~\cite{mcnally2024data}. 
Consequently, compared with directly constructing a downscaling model of ERA5 maps, incorporating satellite observations as conditional inputs via cross-attention module during the downscaling process is essential to ensure that the generated high-resolution ERA5 meteorological data aligns more accurately with actual conditions~\cite{huang2019ccnet, feng2022training, qi2023fatezero}. 


To this end, we propose the \textbf{S}atellite-observations \textbf{G}uided \textbf{D}iffusion Model (\textbf{SGD}) based on the conditional diffusion model. 
As shown in \cref{fig:teaser} \textbf{(b)}, the conditional diffusion model is utilized for the conditional generation of high-resolution ERA5 atmospheric data through the integration of brightness temperature information from GridSat satellite observations, thereby aligning more closely with real-world conditions. 
During the training process, a satellite encoder is pre-trained to extract features from the GridSat maps, which are then fused with ERA5 maps through a cross-attention module. 
The conditional denoising function is trained by the UNet module of the diffusion model. 
During the sampling process, instead of directly modeling the downscaling process, we leveraged its inverse to guide the generation of high-resolution ERA5 maps by incorporating guidance from low-resolution maps and observations from weather stations, which enables SGD to yield high-quality small-scale maps with faithful details.
A convolutional kernel $\mathcal{D}^t_{\varphi}$ with optimizable parameters ${\varphi}$ is utilized to simulate the upscaling process, which is updated in real-time across reverse steps $t$, aiming to guide the details of generated small-scale maps closer to both the original ERA5 maps and station-level observations. 
A distance function is proposed to measure the discrepancy between the two, and its gradient was used to update the mean of the samples, guiding the generation of small-scale maps with fine details based on both the low-resolution maps and station-level observations. 
The convolutional kernel parameters are updated through the gradient with respect to the parameters themselves, ensuring that the model dynamically refines its ability to construct the resolution conversion process. 
Extensive experiments have demonstrated that SGD can generate more accurate high-resolution ERA5 maps than off-the-shelf interpolation-based and diffusion-based methods. 
Ablation studies further validate the effectiveness of utilizing GridSat maps as conditioned inputs. 
Simultaneously, we have also analyzed further optimization solutions for the performance and generation accuracy of SGD through comprehensive experiments. 
Our contributions are three folds:
\begin{itemize}
\item 
We propose SGD for ERA5 meteorological states downscaling. 
By integrating GridSat satellite observations into the conditional diffusion model to capture the coupling between satellite observations and ERA5 maps, SGD is capable of generating atmospheric states that are more aligned with real-world conditions. 
\item
We employ an optimizable convolutional kernel to simulate the upscaling during the sampling process. 
By drawing upon a distance function, we incorporate guidance from both low-resolution ERA5 maps and station-level observations at a scale of $25km$ into the generation of small-scale ERA5 maps, thereby enabling the model to produce high-quality ERA5 maps at $6.25km$ or even smaller scale that exhibit faithful details. 
\end{itemize}

\begin{figure*}[t]
    \centering   
    \includegraphics[width=\linewidth]{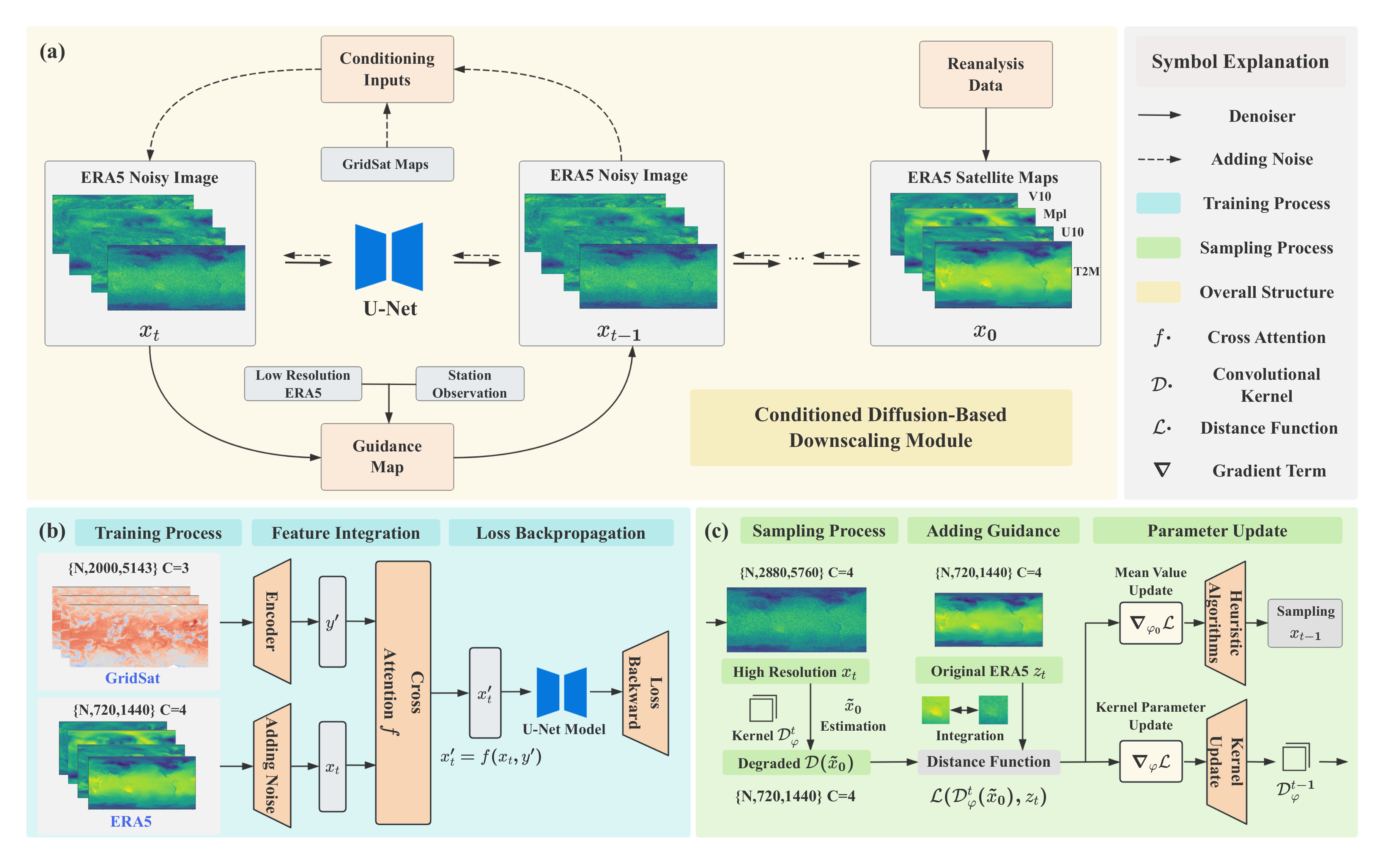}
    \vspace{-1.1cm}
    \caption{
\textbf{(a)} Overview of the conditional diffusion-based downscaling model. 
\textbf{(b)} The satellite observations, GridSat, undergo a feature extraction encoder to serve as the conditional input.
GridSat is then fused with ERA5 via cross-attention to train the conditional diffusion model.
\textbf{(c)} During the sampling process, low-resolution ERA5 maps are utilized to guide the generation of high-resolution maps. 
This is achieved through convolutional kernels $\mathcal{D}$ with optimizable parameters $\varphi$ that facilitate resolution transformation, while a distance function $\mathcal{L}$ is introduced to quantify the disparity between the upscaled convolution-generated map $\mathcal{D}(\tilde{x}_0)$ and the original ERA5 map $z_t$, where $\tilde{x}_0$ refers to the real-time estimation of the generated maps. 
The gradient of the distance function with respect to $\tilde{x}_0$ is utilized to update the mean value used in sampling. 
Simultaneously, the gradient of the distance function concerning the convolutional kernel parameters is employed to update these parameters, thereby enabling a more accurate simulation of the inverse process of downscaling.
}
    \label{fig:main_fig}
\vspace{-0.5cm}
\end{figure*}

%% file: sec/2_related.tex
\section{Related Workds}
\label{sec:related}

\textbf{Downscaling.}
Downscaling aims to transform original low-resolution maps into their smaller-scale counterparts to accurately obtain surface meteorological states at minor locations~\cite{chajaei2024machine, doll2024streamflow, wang2024spatial}. 
Based on this objective, various deep learning methods~\cite{menon2020pulse,maraun2016bias} have been employed for the downscaling of meteorological maps. 
For instance, generative adversarial network (GAN) based methods incorporate a generative adversarial loss which enables the generator to produce results indistinguishable from real maps by the discriminator~\cite{chen2020adversarial}. 
CliGAN~\cite{chaudhuri2020cligan} is a GAN-based downscaling approach for precipitation data derived from global climate models to regional-level gridded data by employing a convolutional encoder-dense decoder network.

In recent years, diffusion models have been widely utilized due to their diversity and their capability to produce high-quality downscaling maps~\cite{fei2023generative}. 
Recent investigation~\cite{bischoff2024unpaired} has centered on the downscaling of fluid flows utilizing generative models founded on diffusion maps and Latent Diffusion Models (LDM). 
Moreover,~\cite{lopez2024dynamical} integrates dynamical downscaling with generative methods to augment the uncertainty estimates of downscaled climate projections, thereby demonstrating the potential of diffusion models in refining downscaling methodologies.
However, these methods overlook the coupling relationship that exists between ERA5 reanalysis data and satellite observations.
Therefore we aim to employ GridSat maps as conditions and downscale the resolution of ERA5 maps from $25km$ to $6.25km$ based on a conditional diffusion model. 

\textbf{Diffusion-based SR.}
Super-resolution is an image processing technique that aims to increase the resolution of the image and enhance the clarity and intricate details within the image~\cite{tu2024taming}. 
Recent advancements in super resolution~\cite{li2022srdiff,liu2022diffusion,gao2023implicit} have seen the emergence of diffusion models as a promising approach to address various challenges in this task, such as over-smoothing, mode collapse, and computational inefficiency. 
~\cite{ho2022cascaded} illustrates the efficacy of cascaded diffusion models in producing high-fidelity images, with a particular emphasis on class-conditional ImageNet generation.
Furthermore,~\cite{liu2024patchscaler} presented PatchScaler, a patch-independent diffusion-based method for single image super-resolution. 
Nonetheless, meteorological data owns its characteristics such as uneven spatial distributions and intricate spatial correlations compared with natural images.
Therefore, we aim to train a meteorology-specific diffusion model that is conditioned on satellite observations, which is able to generate high-resolution meteorological data under the guidance of current low-resolution reanalysis data and station-level observations.

%% file: sec/3_method.tex
\section{Methodology}
The key idea of our satellite observations guided conditional diffusion model, SGD, is to employ satellite observations (GridSat) as conditional inputs to enhance the coherence of the SR maps generated by the conditional diffusion model.
Satellite observations are utilized to obtain the reanalysis data, such as ERA5. 
Therefore, leveraging original satellite observations enables the generated SR maps to align more closely with actual conditions. 
Additionally, LR ERA5 maps serve as the guided maps during the sampling process, ensuring the generation of maps replete with faithful details. 
An overview of our SGD is depicted in ~\cref{fig:main_fig}. 

\subsection{Pre-trained Encoder}
The pre-trained encoder is employed to obtain the embedding feature of each channel within every GridSat map. 
The encoder extracts features through convolutional operations and skips connections, while the decoder reconstructs the features of GridSat via a symmetric architecture. 
The detailed structure of the encoder is shown in Appendix. 
Following the pre-training of the encoder architecture $f$ with parameter $\phi$, SGD is capable of more effectively integrating the features of GridSat maps with those of ERA5 and subsequently inputting them into the UNet-based conditional diffusion model. 
Further analysis of the encoder's role will be conducted in ~\cref{ablation_study}. 

\subsection{Satellite Conditioning Mechanisms}
Traditional unconditional DDPMs do not rely on additional conditions during training, resulting in generated images that tend to lack specific guidance and semantic coherence. 
In contrast, conditional DDPMs guide the generation process by incorporating external conditional information such as labels, textual descriptions, or images, allowing for effective control over the features and attributes of the generated results.
For the downscaling task of ERA5 maps, considering that the ERA5 data is derived from satellite observation, a coupling relationship exists between the two maps. 
The brightness temperature data obtained from satellite observations significantly influence the meteorological states within ERA5.

Therefore, SGD employs a GridSat map conditional DDPM. 
By leveraging the GridSat map as a condition, the diffusion model is transformed into a conditional generator, effectively utilizing the features within GridSat to generate high-resolution atmospheric states that are more consistent with reality during the sampling process. 
After the GridSat features are extracted by the encoder into a latent-space representation, it is then integrated with the ERA5 maps through the cross-attention mechanism to facilitate the fusion of features.
\begin{align}
Atten(Q,K,V)=softmax(\frac{W_Q(x)W_K(y')^T}{\sqrt{d}})\cdot W_V(y').
\end{align}
The matrices $W_Q\in \mathbb{R}^{d\times d_{\epsilon}}, W_K, W_V\in \mathbb{R}^{d\times d_{\tau}}$, which serve as learnable projection matrices with trainable parameters, are respectively utilized to amalgamate the feature information from the ERA5 map $x$ and the conditioning input $y'$ within the same time period. 
$W_Q$ extracts features from ERA5 maps, while $W_K$ and $W_V$ process features derived from GridSat maps. 
The parameters of cross attention module are jointly optimized with the UNet backbone.

Based on this foundation, the map-conditioned DDPM can undergo training according to the following objective.
\begin{align}
E_{\epsilon \sim \mathcal{N}(0,I),y',t\sim [0,T]}[\left \|\epsilon - \epsilon_\theta(x_t,y',t) \right \|^2_2 ].
\end{align}

\subsection{Downscaling ERA5 via Zero-shot Sampling}
In this section, we employ pre-trained GridSat map conditional DDPMs to generate high-resolution ERA5 maps that are faithful to the real surface meteorological conditions. 
Off-the-shelf deep learning methods commonly downscale ERA5 maps without leveraging satellite data, thereby neglecting the intricate coupling between satellite observations and ERA5 maps. 
This overlook will lead to discrepancies between the downscaled outcomes and actual conditions.
Meanwhile, the high-resolution ERA5 map with faithful details contains more precise information and holds greater value in practical data applications as it is more closely aligned with actual observations. 
In pursuit of this effect, SGD utilizes low-resolution ERA5 maps to guide the reverse steps of the conditional diffusion model.
A convolutional kernel with optimizable parameters is developed to achieve a downscaling process at arbitrary resolution. 
The sampling process is capable of downscaling the resolution of ERA5 maps from a scale of 25km to 6.25km with accurate details.

Specifically, as illustrated in \cref{fig:main_fig}, SGD utilizes optimizable convolutional kernels $\mathcal{D}$ with parameter $\varphi$ in the reverse steps of conditional DDPMs to facilitate the resolution transformation of the sampled maps $x_t$.
A distance function $\mathcal{L}$ is then employed to quantify the discrepancy between the convolved sampled map $\mathcal{D}(\tilde{x}_0)$ and the low-resolution ERA5 map $z$. 
The distance function can utilize the mean squared error (MSE) loss between the two maps as guidance from ERA5, or alternatively, integrate station-level observations to facilitate bias correction. 
The latter employs average mean absolute error (MAE) loss across two maps for each variable at global stations, thereby ensuring that SGD yields more precise numerical results at each station. 
Specific configurations and the effects of the distance function are shown in ~\cref{distance_function}. 
The gradients of distance function $\mathcal{L}$ are calculated to update the mean $\mu$ and variance $\Sigma$ utilized for sampling $x_{t-1}$, thereby ensuring the generated high-resolution images possess more accurate and rich details which are consistent with ERA5 map at a scale of $25km$. 
Concurrently, the gradients of $\mathcal{L}$ with respect to the convolutional kernel parameters $\varphi$ are also utilized to update themselves. 
This dynamic update mechanism allows the convolutional kernels to simulate the reverse of downscaling more accurately in the sampling process. 

Detailedly, $x_{t-1}$ is sampled by the conditional distribution $p(x_{t-1}|x_{t},y^\prime,z)$, where $y^\prime$ refers to the embedding representation of the GirdSat maps after the encoder module. 
Previous studies~\citep{dhariwal2021diffusion} have derived the conditional transformation formula of the sampling process: 
\begin{align}
\log_{}{p_\theta(x_{t}|x_{t+1},y^\prime,z)}&=\log_{}{(p_\theta(x_{t}|x_{t+1},y^\prime)p(z|x_{t})) }+N_1 \\
&\approx \log_{}{p_\theta(w)+N_2},
\end{align}
where $w\sim \mathcal{N}(w;\mu_\theta(x_t,y^\prime,t)+\Sigma \nabla_{x_t}\log_{}{p(z|x_t)}|_{x_t=\mu},\Sigma I)$, $N_1$ and $N_2$ is constants. The mean $\mu=\mu_\theta(x_t,y^\prime,t)$ and variance $\Sigma=\Sigma(x_t,y^\prime,t)$ is obtained by the conditonal DDPMs. 

Therefore, the sampling distributions integrate the conditional DDPMs with the gradient term, controlling the generation of high-resolution maps.
Inspired by~\cite{fei2023generative}, we employ a heuristic algorithm to approximate the value of the gradient term:
\begin{align}
\log_{}{ p(z\mid x_{t})}&=- \log_{}{N}-s\mathcal{L}( \mathcal{D}(\tilde{x}_0),z)\\
\nabla _{x_t}\log_{}{p(z|x_t)}&=-s\nabla_{\tilde{x}_0}\mathcal{L}(\mathcal{D}(\tilde{x}_0),z),
\end{align}
where $s$ refers to the parameter of the guidance scale which serves as a weight to control the guidance degree.
The gradient of the distance function with respect to $\tilde{x}_0$ is utilized for the calculation of the gradient term, which ultimately updates the mean of the reverse process. 

Based on this, we can derive the complete algorithm for the sampling process.
As shown in \Cref{alg.2}, our model undergoes $T$ reverse steps to sample the high-resolution ERA5 map $x_0$. 
At each step, the optimizable convolutional kernel up-scales the instantaneous estimated value $\tilde{x}_0$ to establish a distance function with the low resolution map $z$.
The gradients of distance function $\mathcal{L}$ with respect to $\tilde{x}_0$ and the parameters of convolution kernel are used to update the mean $\tilde{\mu}_t$ and the convolution kernel parameters, respectively. 
The real-time update mechanism ensures the accuracy of the model's sampling process, making the guidance of low-resolution ERA5 maps and station-level observations more precise and effective. 

Due to the significantly higher map resolution in the generation space compared to that of the pre-trained conditional DDPM, a patch-based approach has been employed to tackle this issue. 
The patch-based method partitions the high-resolution ERA5 map into multiple sub-regions based on a fixed stride, and calculates the corresponding distance metrics and gradient terms for each subregion with the corresponding segments of the guided map $z$. 
The average of the gradient terms across all sub-regions is then utilized for the update of the mean of sampling steps and the convolution kernel parameters. 
The detailed algorithmic framework of the patch-based methodology is introduced in Appendix. 

\begin{algorithm}[t]\small
\renewcommand{\algorithmicrequire}{\textbf{Input:}}
\renewcommand{\algorithmicensure}{\textbf{Output:}}
\caption{\textbf{Sampling Process:} Guided diffusion model with the guidance of low-resolution ERA5 map $z$. Given a conditional diffusion model pre-trained on ERA5 and GridSat maps $\epsilon_\theta(x_t,y,t)$. }
\label{alg.2}
\begin{algorithmic}[1]
\REQUIRE Conditioning input GridSat satellite observation map $y$, low-resolution ERA5 map $z$. Downscaling convolutional function kernel $\mathcal{D}$ with parameter $\varphi$. Pre-trained encoder module $f$ with parameter $\phi$. Learning late $l$ and guidance scale $s$. Distance measure function $\mathcal{L}$. 
\ENSURE Output high-resolution ERA5 map $x_0$. 

\STATE Sample $x_T$ from $\mathcal{N}(0,I)$

\STATE $y' \gets f_\phi(y)$

    \FORALL{t from T to 1}
        \STATE $\tilde{x} _0=\frac{x_t}{\sqrt{\bar{\alpha}_t }}-\frac{\sqrt{1-\bar{\alpha}_t}\epsilon_\theta(x_t,y',t)}{\sqrt{\bar{\alpha}_t }}$\
        
        \STATE $\mathcal{L}_{\varphi,\tilde{x}_0} =\mathcal{L}(z,\mathcal{D}^{\varphi}(\tilde{x}_0))$\

        \STATE $\tilde{x}_0 \gets \tilde{x}_0-\frac{s(1-\bar{\alpha}_t) }{\sqrt{\bar{\alpha}_{t-1}}\beta_t}\nabla_{{\tilde{x}}_0}\mathcal{L}_{\varphi,\tilde{x}_0}$\
        \STATE $\tilde{\mu}_t=\frac{\sqrt{\bar{\alpha}_{t-1}}\beta_t}{1-\bar{\alpha}_t}\tilde{x}_0+\frac{\sqrt{\bar{\alpha}_{t}}(1-\bar{\alpha}_{t-1})}{1-\bar{\alpha}_t}{x}_t$\

        \STATE $\tilde{\beta}_t=\frac{1-\bar{\alpha}_{t-1}}{1-\bar{\alpha}_t}\beta_t$\
        
        \STATE Sample $x_{t-1}$ from $\mathcal{N}(\tilde{\mu}_t,\tilde{\beta}_tI)$\

        \STATE $\varphi \gets \varphi-l\nabla_{\varphi}\mathcal{L}_{\varphi,\tilde{x}_0}$\
    \ENDFOR
\STATE \textbf{return}  $x_0$
    \end{algorithmic}
\end{algorithm}

%% file: sec/4_experiment.tex
\section{Experiments}

\subsection{Datasets}
\textbf{ERA5.} ERA5~\cite{hersbach2020era5} is a global meteorological reanalysis dataset provided by the European Centre for Medium-Range Weather Forecasting (ECMWF), encompassing data since 1979. 
Its atmospheric variables are derived from atmospheric data collected at 37 different altitude levels, and it also includes several variables that represent Earth surface meteorological information. 
The entire grid comprises 721 latitude and 1440 longitude grid points. 
In our experiment, we selected four surface-level variables for the global: $U_{10}$, $V_{10}$, $T_{2M}$, and $MSL$. 
All four variables exhibit a correlation with human activities.
The specific meanings and units of these variables are outlined in \cref{tab:dataset}. 

\noindent\textbf{GridSat.} GridSat~\cite{skofronick2015global} dataset is a comprehensive series of satellite data products provided by the National Environmental Satellite, Data, and Information Service (NESDIS) of the National Oceanic and Atmospheric Administration (NOAA).  
GridSat dataset is primarily based on Earth observation data acquired from NOAA's geostationary satellite. 
These data within the GridSat collection are routinely employed in meteorological analyses and climatological studies for the estimation of cloud cover and surface temperature, thereby contributing to a more accurate representation of atmospheric variables. 
Among the multitude of variables within this dataset, we have selected three specific parameters: IrWin\_CDR, Irwin\_VZA\_Adj, and IrWVP. 
The detailed information of these variables are illustrated in \cref{tab:dataset}.

\begin{table}[t]
\centering
\caption{Variable information of ERA5 and GridSat datasets.}
\vspace{-0.3cm}
\resizebox{\linewidth}{!}{
\begin{tabular}{c|c c c c}
    \toprule[1pt]
     Dataset&Variable&Abbrev.&Unit&ECMWF\;ID\\
    \midrule
    \multirow{4}{*}{ERA5}
    &U-component of Wind at 10m&$U_{10}$&$m/s$&165\\
    &V-component of Wind at 10m&$V_{10}$&$m/s$&166\\
    &2-Meter Temperature&$T_{2M}$&$K$&167\\
    &Mean Sea Level Pressure&$MSL$&$10^2Pa$&151\\
    \midrule
    \multirow{3}{*}{GridSat}
    &Infrared Window Channel Data Record&IrWin\_CDR&$K$&-\\
    &Viewing Zenith Angle Adjusted&Irwin\_VZA\_Adj&$K$&-\\
    &Integrated Water Vapor Path&IrWVP&kg/$m^2$&-\\
    \bottomrule[1pt]

  \end{tabular}
  }
 \label{tab:dataset}
\vspace{-0.4cm}
\end{table}

\begin{figure*}[t]
    \centering
\includegraphics[width=\linewidth]{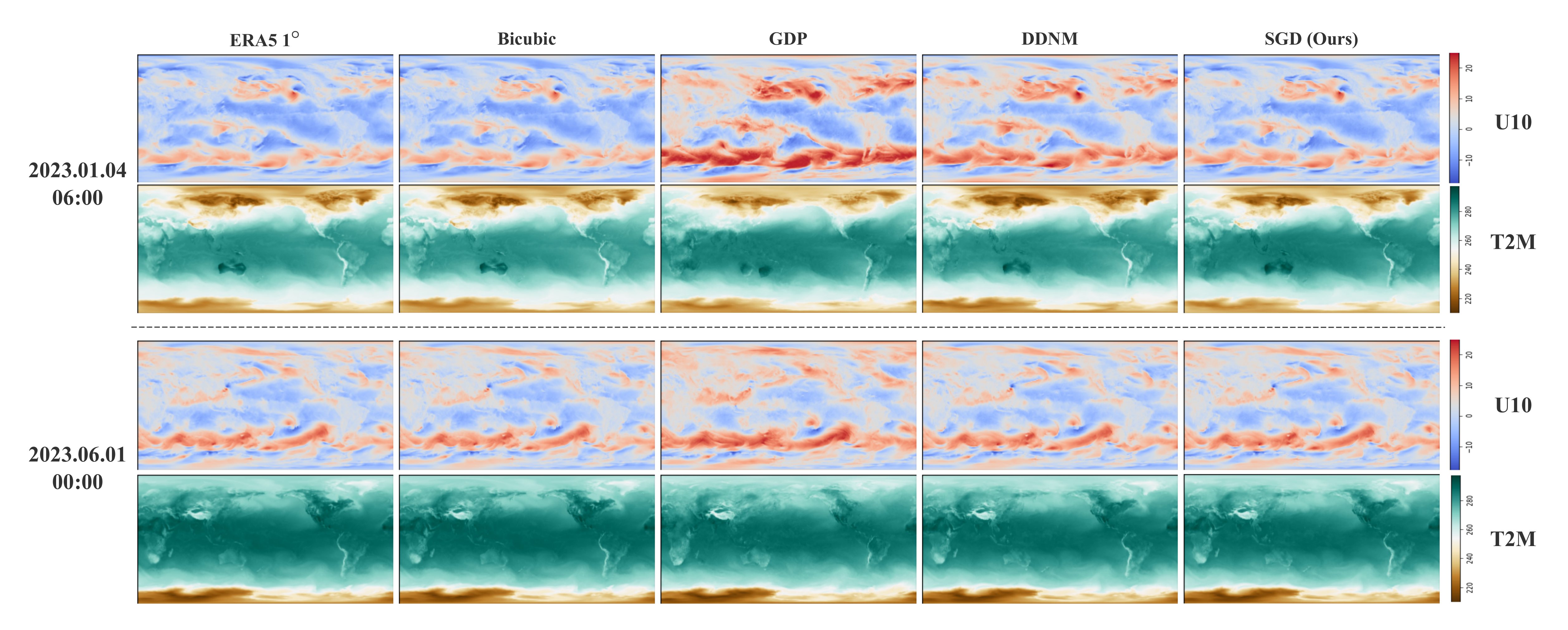}
\vspace{-1cm}
    \caption{
Visualization comparison of different interpolation-based and diffusion-based downscaling results in various time stamps. 
We use different colors to distinguish various variables. 
Our SGD generates downscaling ERA5 maps with faithful details from ERA5 $1^{\circ}$. }
    \label{fig:comparison}
\vspace{-0.4cm}
\end{figure*}

\begin{table}[t]\small
  \centering
  \caption{Station-level downscaling results for $U_{10}$, $V_{10}$, $T_{2M}$ and $MSL$ of various methods. }
  \vspace{-0.3cm}
  \label{tab:main}
  \resizebox{\linewidth}{!}{
  \begin{tabular}{c|c c |c c |c c |c c }
    \toprule[1pt]
     \multirow{2}{*}{Methods}&\multicolumn{2}{c|}{$U_{10}$}&\multicolumn{2}{c|}{$V_{10}$}&\multicolumn{2}{c|}{$T_{2m}$}&\multicolumn{2}{c}{$MSL$}\\
     \cmidrule(lr){2-9}
     
    &MSE&MAE&MSE&MAE&MSE&MAE&MSE&MAE\\
    \midrule
ERA5 $1^{\circ} $  & 53.18 & 5.95 & 38.51 & \textbf{4.95} & 216.27 & 11.39 & 470.06 & 15.78 \\
Bilinear  & 56.55 & 6.16 & 37.94 & 5.03 & 198.40 & 11.26 & 464.51 & 16.05 \\
Bicubic  & 55.74 & 6.11 & \textbf{37.88} & 4.98 & 201.03 & 11.15 & 458.82 & 15.97 \\
DGP~\cite{pan2021exploiting}  & 97.02 & 7.96 & 47.52 & 5.07 & 214.58 & 11.94 & 529.72 & 18.54 \\
GDP~\cite{fei2023generative}  & 94.99 & 7.85 & 40.17 & 5.04 & 190.11 & 10.82 & 511.71 & 18.14 \\
DDNM~\cite{wang2022zero}  & 52.08 & 5.94 & 42.17 & 5.65 & 193.24 & 11.77 & 396.58 & 15.12 \\
SwinRDM~\cite{chen2023swinrdm}& 53.18 & 5.95 & \textbf{38.51} & \textbf{4.95} & 216.27 & 11.39 & 470.06 & 15.78 \\
Ref-SR~\cite{huang2022task}& 62.72 & 6.15 & 43.12 & 5.17 & 195.42 & 11.02 & 395.42 & 15.10 \\
$C^2$-Matching~\cite{jiang2021robust}& 65.12 & 6.02 & 44.57 & 5.41 & 200.17 & 11.36 & 410.72 & 15.32 \\
HyperDS~\cite{liu2024observation}& 53.72 & 6.01 & 41.37 & 5.26 & 191.83 & 10.87 & 384.72 & 14.72 \\
\midrule
SGD& \textbf{51.65} & \textbf{5.84} & 39.82 & 5.05 & \textbf{187.69} & \textbf{10.63} & \textbf{374.39} & \textbf{14.49} \\
    \bottomrule[1pt]

  \end{tabular}
  }
\vspace{-0.5cm}
\end{table}

\begin{table*}[t]\small
  \centering
  \caption{Ablation study regarding the effectiveness of the cross attention module (CA) and pre-trained encoder (PE). }
  \vspace{-0.3cm}
  \label{tab:ablation_training}
  \resizebox{0.9\linewidth}{!}{
  \begin{tabular}{c|c c|c c|c c|c c|c c}
    \toprule[1pt]
     \multirow{2}{*}{Methods} &\multicolumn{2}{c}{ Module}&\multicolumn{2}{|c|}{$U_{10}$}&\multicolumn{2}{c|}{$V_{10}$}&\multicolumn{2}{c|}{$T_{2M}$}&\multicolumn{2}{c}{$MSL$}\\
     \cmidrule(lr){2-3}
     \cmidrule(lr){4-11}

    &{CA}&{PE}&{MSE}&{MAE}&{MSE}&{MAE}&{MSE}&{MAE}&{MSE}&{MAE}\\
    \midrule
    {\small Unconditional DDPM Based SGD}&\XSolidBrush&\XSolidBrush&75.3985&6.9324&46.9961&5.4979&210.1230&11.2656&589.2316&18.7626\\
    {\small Conditional DDPM Based SGD}&\Checkmark&\XSolidBrush&58.3419&6.2400&45.2718&5.3794&207.4618&11.2806&397.2393&15.2934\\
    \midrule
    {\small SGD with Pre-trained Encoder}&\Checkmark&\Checkmark&\textbf{51.6512}&\textbf{5.8440}&\textbf{39.8189}&\textbf{5.0479}&\textbf{187.6857}&\textbf{10.6311}&\textbf{373.3909}&\textbf{14.4917}\\
    \bottomrule[1pt]

  \end{tabular}
}
\vspace{-0.3cm}
\end{table*}

\begin{table*}[t]\footnotesize
  \centering
  \caption{Ablation study regarding the downscaling results guided by different distance functions during the sampling process. }
  \vspace{-0.3cm}
  \label{tab:ablation_guidance}
  \resizebox{0.9\linewidth}{!}{
  \begin{tabular}{c|c c|c c|c c|c c|c c}
    \toprule[1pt]
     \multirow{2}{*}{Methods} &\multicolumn{2}{c}{ Guidance}&\multicolumn{2}{|c|}{$U_{10}$}&\multicolumn{2}{c|}{$V_{10}$}&\multicolumn{2}{c|}{$T_{2M}$}&\multicolumn{2}{c}{$MSL$}\\
     \cmidrule(lr){2-3}
     \cmidrule(lr){4-11}

    &{ERA5}&{Station}&{MSE}&{MAE}&{MSE}&{MAE}&{MSE}&{MAE}&{MSE}&{MAE}\\
    \midrule
    {\small ERA5 Guided SGD}&\Checkmark&\XSolidBrush&63.32&6.55&45.84&5.39&198.14&10.91&427.25&15.52\\
    {\small Station Guided SGD}&\XSolidBrush&\Checkmark&\textbf{43.58}&\textbf{5.52}&\textbf{39.21}&\textbf{4.98}&203.51&11.22&415.76&16.57\\
    {\small ERA5 + Station Guided SGD}&\Checkmark&\Checkmark&48.78&5.64&41.43&5.15&\textbf{194.66}&\textbf{10.67}&\textbf{374.56}&\textbf{14.27}\\
    \bottomrule[1pt]

  \end{tabular}
  }
\vspace{-0.5cm}
\end{table*}

\begin{figure}[t]
    \centering
\includegraphics[width=\linewidth]{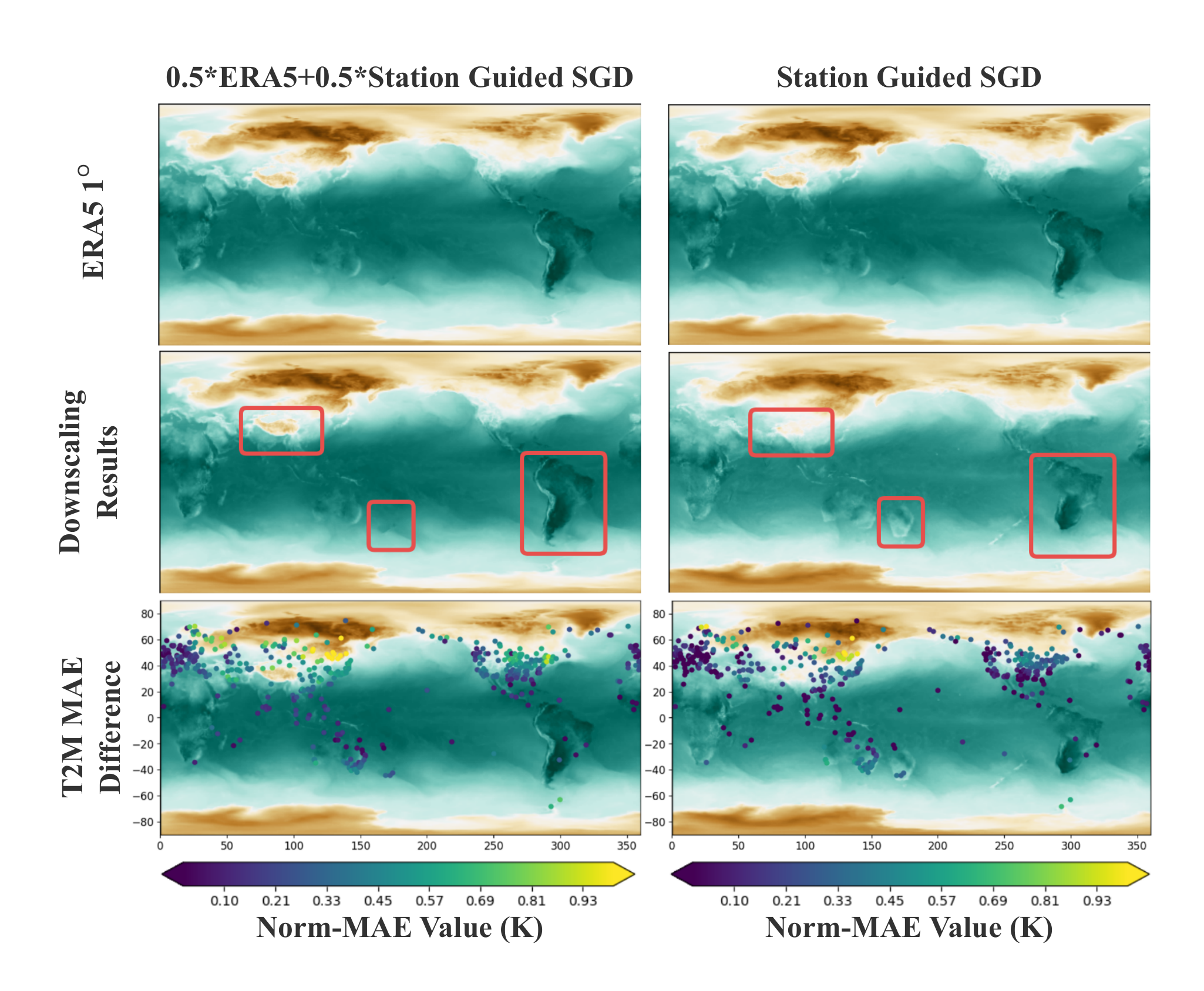}
    \vspace{-1.0cm}
    \caption{
Visualization comparison between totally using station observation as guidance and both integrating ERA5 and station observation as guidance. 
The former has a smaller MAE loss with the station observation, while the latter has more faithful details. 
As for MAE difference, the darker color of the observation station means the MAE bias is smaller. 
}
    \label{fig:guidance}
\vspace{-0.5cm}
\end{figure}

\begin{figure*}[t]
    \centering
\includegraphics[width=\linewidth]{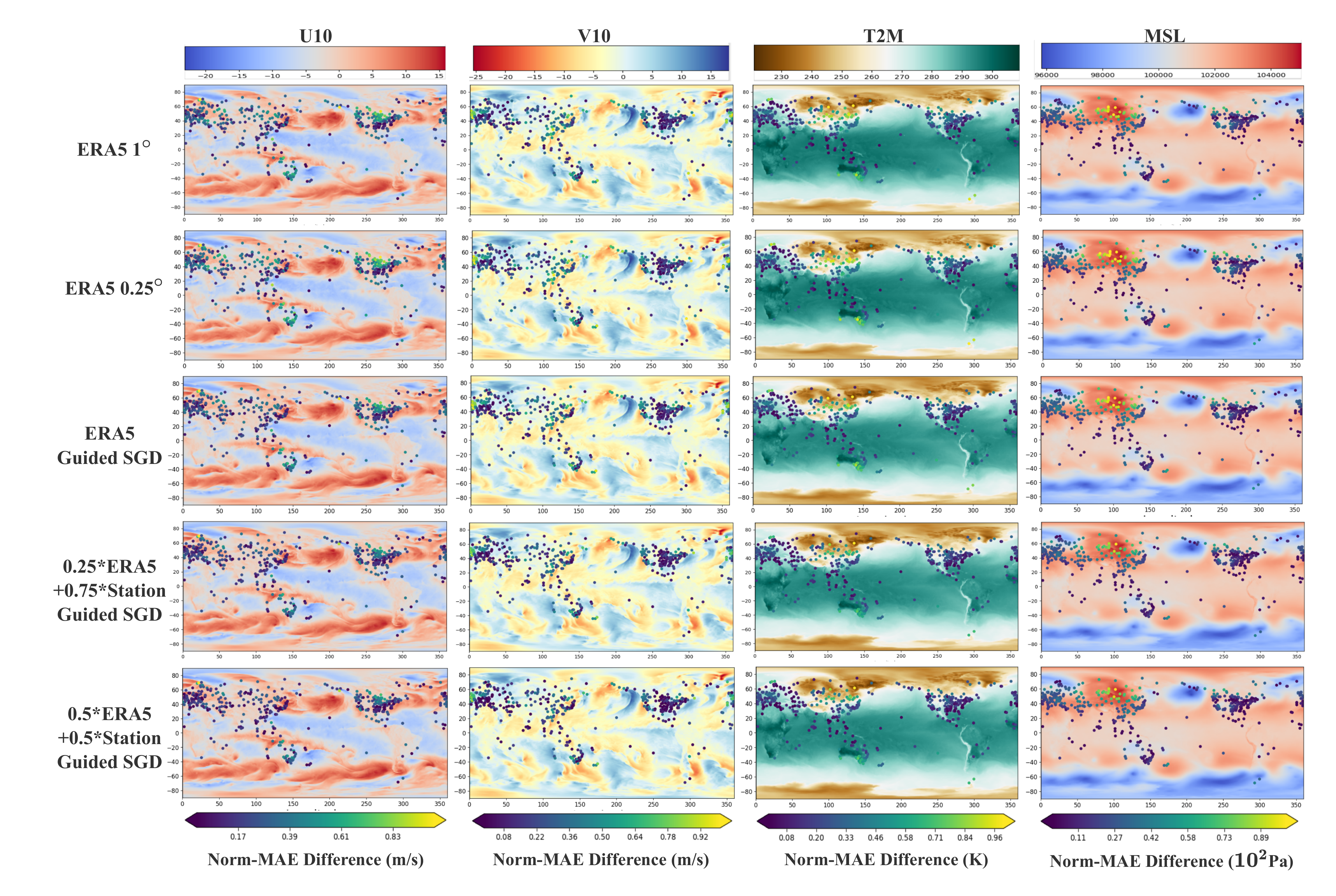}
\vspace{-1cm}
    \caption{Visualization comparison of SGD downscaling to station-scale employing various distance functions, where the coloration of each observation station signifies the MAE loss between the downscaled results and their corresponding observed values. }
    \label{fig:distance_function}
\vspace{-0.5cm}
\end{figure*}

\subsection{Implementation Details}
We have trained SGD on the ERA5 reanalysis dataset and the GridSat satellite observation maps for approximately 200,000 steps. 
The temporal intervals of the samples within both the ERA5 and GridSat datasets are 6 hours. 
We have selected the data from 2010 to 2021 as the training set, the entirety of the 2022 data as the validation set, and the first 6 months of 2023 as the test set.
The ERA5 maps employed in the training encompass four channels, each with a shape of $720\times1440$, while the GridSat maps feature three channels, each channel have a shape of $2000\times5143$. 
All the training task are conducted on NVIDIA A100 80GB GPU. 
For the training process, SGD is optimized using AdamW with $\beta_1=0.9$ and $\beta_2=0.999$ in 16-bit precision with loss scaling, while keeping 32-bit weights, Exponential Moving Average (EMA), and optimizer state. 
The pre-trained encoder module employs the mean squared error as its loss function during training, undergoing the process for 100 epochs. 
During the sampling phase, we utilize the same UNet structure as in the training process of conditional DDPM and employ an identical noise schedule. 
The kernel size used for simulating upscaling is $9\times9$ and the variance $\beta_t$ we utilize undergoes a linear increase from $\beta_1=10^{-4}$ to $\beta_T=0.02$. 

\subsection{Evaluation Metrics}
To compare the performance disparities between the proposed SGD and other interpolation-based and diffusion-based downscaling methodologies, we employ the interpolation of ERA5 reanalysis maps into station scale to validate the efficacy of our model. 
These compared methods leverage interpolation to extract meteorological state variables from ERA5 maps at the scale of observation stations. 
Specifically, we utilize the \textit{grid\_sample} function from the PyTorch to directly interpolate ERA5 maps with resolutions of $1^{\circ}$ and $0.25^{\circ}$, which is based on the absolute latitude and longitude positions of global stations from the Weather5k dataset~\cite{han2024weather}, which is a large-scale time series forecasting dataset containing weather data from 5,672 weather stations worldwide.  
We select MAE loss and MSE loss to quantify the inherent bias between the ERA5 maps derived from various downscaling methods and station observations. 

\subsection{Main Results}
In this section, we conduct qualitative comparisons on downscaled maps with other methods and quantitative comparisons within the inherent bias of generated ERA5 meteorological state variables at the scale of observation stations. 
The interpolation-based methods encompass bilinear and bicubic interpolation, while the diffusion-based methods include GDP~\cite{fei2023generative}, and DDNM~\cite{wang2022zero}. 
As depicted in ~\cref{tab:main}, when performing station-level downscaling evaluation, SGD surpasses existing methods in terms of both MAE and MSE metrics across variables $U_{10}$, $T_{2M}$ and $MSL$. 
Compared to ERA5 $1^{\circ}$, SGD presents a prominent enhancement in the T2M and MSL variables.
This reflects that employing the brightness temperature data from GridSat as conditioned input enables SGD to yield downscaling results that exhibit a smaller bias and greater accuracy in temperature and other variables, which is closely aligned with actual values. 
The downscaling results also show an improvement in metrics compared to ERA5 $1^{\circ}$ and $0.25^{\circ}$. 
The metrics comparison demonstrates that the downscaling results generated by SGD exhibit a reduced bias with the observation values from global stations. 
~\cref{fig:comparison} presents the qualitative comparison of the downscaling results produced by SGD and other methods. 
It is shown that SGD generates ERA5 maps at a scale of $6.25km$ with more faithful details, validating its adeptness in capturing and restoring intricate details during the downscaling process. 

\subsection{Station Observations as Additional Guidance}
\label{distance_function}
Within the sampling process, SGD incorporates ERA5 maps as a guiding framework, thereby enabling the generation of downscaling results containing faithful details. 
The guided sampling is achieved through the utilization of a distance function to evaluate the disparity between the maps upscaled by optimizable convolution kernels and the low-resolution ERA5 maps. 
One feasible approach involves harnessing the MSE loss between the two maps. 
Nevertheless, it is also needed to ensure that the downscaling outcomes of SGD on the Weather5k observations align more closely with the actual atmospheric conditions, 
As illustrated in ~\cref{tab:ablation_guidance}, integrating the MAE loss of the SGD downscaling maps at the observation stations in the distance loss yields superior metrics and reduced disparities across multiple variables in SGD. 

It is noteworthy that employing solely the bias of the observation station as the distance function for SGD can bring the variables at the stations closer to the actual observations. 
However, this approach may introduce discrepancies in local regions at the overall downscaling results when compared to ERA5, as shown in ~\cref{fig:guidance}. 
Therefore, to seek a balance between the faithful details and reducing the observation bias, we incorporated both factors into the distance function and conducted comparative analyses across various weight relationships. 
~\cref{fig:distance_function} presents the SGD downscaling results under various settings of the distance function, as well as the inherent biases in meteorological state variables within ERA5 maps as observed at various observation stations. 
As depicted in ~\cref{fig:distance_function}, the incorporation of the observation station loss within the distance function facilitates the generation of high-resolution ERA5 maps that exhibit a smaller MSE loss relative to the true observational results at most stations. 

\subsection{Ablation Study}
\label{ablation_study}
\textbf{The Effectiveness of the Conditional GridSat Maps.} 
To assess the effectiveness of incorporating GridSat satellite observations on the diffusion model of SGD, we conducted an ablation study comparing the model to its unconditional counterpart. 
The unconditional diffusion model solely incorporates ERA5 maps into its training without GridSat as conditions.
As shown in \cref{tab:ablation_training}, the conditional diffusion model (SGD) demonstrates a superior downscaling performance across all metrics compared to the unconditional diffusion model. 

\noindent \textbf{The Effectiveness of the Pre-trained GridSat Encoder.} 
Before performing feature fusion on GridSat maps, SGD utilizes an encoder to extract its features. 
We conduct experiments on whether this GridSat encoder is pre-trained.
As depicted in \cref{tab:ablation_training}, SGD equipped with the pre-trained encoder showcases enhancements in various variables compared to its counterpart without the pre-trained GridSat encoder. 
This validates the efficacy of the pre-trained encoder in facilitating the SGD's ability to more effectively extract features from GridSat maps.

%% file: sec/5_conclusion.tex
\section{Conclusion}
We propose SGD, a conditional diffusion model for robust downscaling, which enables the downscaling of ERA5 maps to arbitrary resolutions for the extraction of intricate meteorological states. 
Specifically, considering that ERA5 data is derived from satellite observation data and the brightness temperature data from satellite observations significantly influences the meteorological states within ERA5, SGD employs GridSat satellite observation maps as conditions to generate downscaled ERA5 maps that more accurately align with actual meteorological states. 
During the sampling process, SGD utilizes the generative prior within the conditional diffusion model, thereby incorporating guidance both from low-resolution ERA5 maps and station-level observations through the optimizable kernel and distance functions. 
Experiments demonstrate that SGD is capable of generating atmospheric states in high-resolution maps with more ideal accuracy and faithful details than various off-the-shelf methods
SGD also showcases its capability of producing high-quality ERA5 maps at a small scale of $6.25km$. 

\noindent \textbf{Limitations and Future Work.}
The training data in SGD consists of ERA5 and GridSat.
However, SGD serves as a versatile framework that could incorporate more modalities such as other reanalysis data, observations from polar-orbiting satellites, sounding and radar data.
Once these systematic data are all integrated into SGD, more accurate weather conditions near the surface can be achieved.

%% file: sec/X_suppl.tex
\clearpage
\setcounter{page}{1}
\maketitlesupplementary

\input{app/preliminary}

\input{app/patch_based_methods}

\input{app/encoder}

\input{app/more_results}

\input{app/ablation}

\input{app/running_time}

\input{app/ablation_variables}

%% file: app/preliminary.tex
\section{Preliminary}
\label{app:preliminary}
Unconditional diffusion model, proposed by~\cite{ho2020denoising}, is a powerful generative model composed of a forward process and a reverse process. 
The former aims to gradually introduce random Gaussian noise into the original images over $T$ diffusion steps, ultimately resulting in pure Gaussian noise $x_T\sim \mathcal{N}(0,I)$. 
The latter, being the reverse of the forward process, intends to denoise and sample the generated images from the pure Gaussian noise through a pre-trained noise estimation network. 

The forward process is a Markov chain without learnable parameters. 
The denoising method for each step is defined by the following equation, where $\beta_t$ refers to the variance of the forward process, which is experimentally set as a hyperparameter solely dependent on the diffusion steps $t$. 
\begin{align}
q (x_t|x_{t-1})=\mathcal{N}(x_t;\sqrt{1-\beta_t}x_{t-1},\beta_tI).
\end{align}

For each steps in the reverse process $p(x_{t-1}|x_t)=\mathcal{N}(x_{t-1};\mu_\theta(x_t,t),\Sigma_\theta I)$, the mean of the distribution is hard to compute directly as the forward process. 
Consequently, we necessitate the utilization of a neural network with parameter $\theta$ to estimate the noise inherent within the image $x_t$. 
By employing Bayes theorem, we can express the mean and variance of the reverse process as follows:
\begin{align}
\mu_\theta(x_t,t)&=\frac{1}{\sqrt{\alpha_t}}(x_t-\frac{\beta_t}{\sqrt{1-\bar{\alpha}_t}}\epsilon_\theta(x_t,t))\\
\Sigma_\theta(x_t)&=\frac{1-\bar{\alpha}_{t-1} }{1-\bar{\alpha}_{t}}\beta_t,
\end{align}

Among them, $\epsilon_\theta(x_t,t)$ represents the noise estimation function, which is pre-trained by utilizing the low-resolution ERA5 maps. 
It performs real-time estimation and simulation of the noise contained within the maps, thereby enabling denoising to sample $x_{t-1}$. 
The unconditional diffusion model is trained utilizing maximum likelihood estimation, with the objective for each training iteration defined as follows: 
\begin{align}
E_{\epsilon \sim \mathcal{N}(0,I),t\sim [0,T]}[\left \|\epsilon - \epsilon_\theta(x_t,t) \right \|^2 ].
\end{align}

%% file: app/patch_based_methods.tex
\section{Patch-based Methods}
The scale of the ERA5 maps used as input for SGD reaches $25km\times 25km$. 
To address the downscaling task at this resolution, we employed a patch-based method during the sampling process of the conditional DDPM. 
The detailed introduction of this method is shown in \cref{app_alg}, the patch-based method partitions the low-resolution ERA5 maps into several sub-regions based on a fixed stride and size. 
For each individual sub-region, the gradient term of the distance loss between the ERA5 maps obtained from convolution kernels and the corresponding low-resolution sub-region map is computed separately. 
Subsequently, the mean of the Gaussian distribution and the parameters of the convolution kernels in each sub-region are updated based on these gradient terms. 
The overall map is then updated by averaging the updated values across all sub-regions, each weighted by a binary patch mask that quantifies the regional scope, thereby refining the overall mean and convolution kernel parameters of the sampled high-resolution ERA5 map, resulting in smoother generated maps. 
By leveraging this strategy, SGD is capable of downscaling ERA5 maps to any desired resolution, further enhancing the practicality of the model. 

\begin{algorithm}[t]\small
\renewcommand{\algorithmicrequire}{\textbf{Input:}}
\renewcommand{\algorithmicensure}{\textbf{Output:}}
\caption{Patch-based Methods of SGD: Guided diffusion model with the guidance of low-resolution ERA5 map $z$. Given a conditional diffusion model pre-trained on ERA5 and GridSat maps $\epsilon_\theta(x_t,y,t)$. }
\label{app_alg}
\begin{algorithmic}[1]
\REQUIRE Conditional input GridSat satellite observation map $y$, low-resolution ERA5 map $z$. Downscaling convolutional kernel $\mathcal{D}$ with parameter $\varphi$. Pre-trained encoder module $f$ with parameter $\phi$. Learning rate $l$ and guidance scale $s$. Distance measure function $\mathcal{L}$. Overlapping patch stride $r$, overlapping patch size $v=720\times 1440$. Overlapping patch set $K$, each patch commences its traversal from the top-left block of the $720\times 1440$ grid on the maps, progressing sequentially with a displacement of stride $r$. A binary patch mask set $\left \{ P^k \right \} ,k\in K$.
\ENSURE Output high resolution ERA5 map $x_0$.
\STATE Sample $x_T$ from $\mathcal{N}(0,I)$ 
\STATE $y'=f_\phi(y)$
    \FORALL{t from T to 1}
    
        \STATE $\tilde{x} _0=\frac{x_t}{\sqrt{\bar{\alpha}_t }}-\frac{\sqrt{1-\bar{\alpha}_t}\epsilon_\theta(x_t,t)}{\sqrt{\bar{\alpha}_t }}$\

        \FORALL{i from 1 to $\left | K \right | $}
        \STATE $\mathcal{L}_{\varphi^i,\tilde{x}_0} =\mathcal{L}(z\circ P^i,\mathcal{D}^{\varphi^i}(\tilde{x}_0\circ P^i))$\

        \STATE $\varphi^i \gets \varphi^i-l\nabla_{\varphi^i}\mathcal{L}_{\varphi^i,\tilde{x}_0}$\

        \STATE $\tilde{x}_0^i \gets \tilde{x}_0^i-\frac{s(1-\bar{\alpha}_t) }{\sqrt{\bar{\alpha}_{t-1}}\beta_t}\nabla_{{\tilde{x}}_0}\mathcal{L}_{\varphi^i,\tilde{x}_0}$\

        \STATE $\tilde{\mu}_t^i=\frac{\sqrt{\bar{\alpha}_{t-1}}\beta_t}{1-\bar{\alpha}_t}\tilde{x}_0^i+\frac{\sqrt{\bar{\alpha}_{t}}(1-\bar{\alpha}_{t-1})}{1-\bar{\alpha}_t}{x}_t$\

        \ENDFOR
        
        \STATE $\varphi =\frac{1}{\left | K \right | }\sum_{j=1}^{\left | K \right | } \varphi^j \circ P^j$\
        
       \STATE $\tilde{\mu}_t=\frac{1}{\left | K \right | }\sum_{j=1}^{\left | K \right | } \tilde{\mu}_t^j \circ P^j$\

        \STATE $\tilde{\beta}_t=\frac{1-\bar{\alpha}_{t-1}}{1-\bar{\alpha}_t}\beta_t$\
        \ENDFOR
        
        \STATE Sample $x_{t-1}$ from $\mathcal{N}(\tilde{\mu}_t,\tilde{\beta}_tI)$\
    
\textbf{return} $x_0$
    \end{algorithmic}
\end{algorithm}

%% file: app/encoder.tex
\section{Pre-trained Encoder}
\begin{figure}[t]
    \centering
\includegraphics[width=\linewidth]{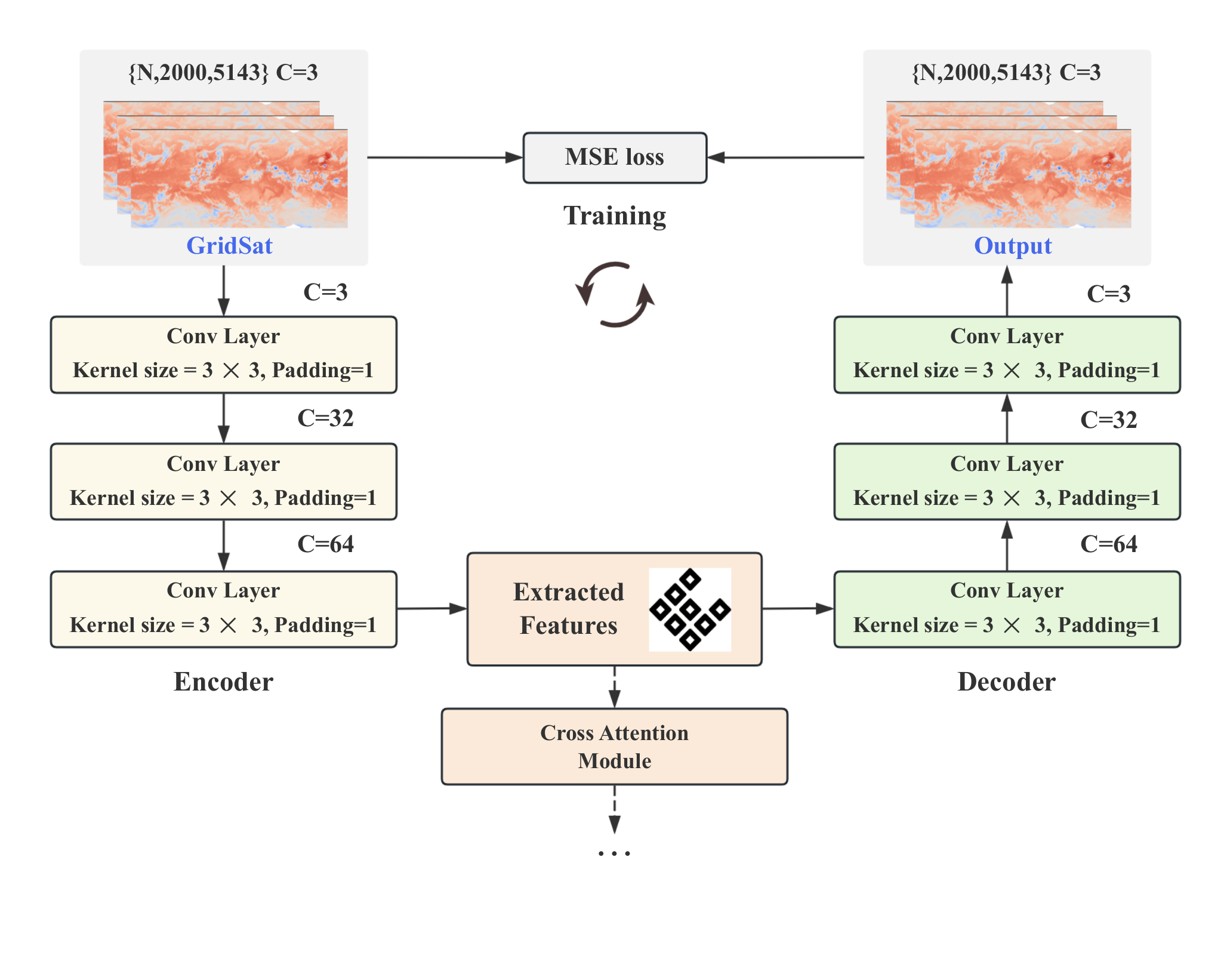}
    \caption{Overall architecture of the encoder and decoder modules used in SGD. }
    \label{fig:app_feature}
\end{figure}

Before utilizing cross attention for feature fusion, SGD necessitates the extraction of features from GridSat maps by an encoder. 
The pre-trained encoder aims to enhance the feature extraction capabilities of SGD and its downscaling performance. 
The pre-trained module comprises two components: the encoder and a decoder of symmetric structure. 
The former is utilize to extracte features from GridSat maps into latent space, while the latter aims to reconstruct the encoder's outputs. 
The encoder module consists of several convolutional layers, employing $3\times 3$ convolutional kernels with a padding of 1, elevating the GridSat maps' channel count to 64. Similarly, the decoder also encompasses convolutional layers, responsible for the reconstruction of the extracted features, the detailed structure is shown in \cref{fig:app_feature}. 
The training objective is to minimize the MSE loss between the input GridSat maps and the output maps post-decoder, with the total training epochs approximating 100. 

%% file: app/more_results.tex
\section{Additional Visualization Results}
\begin{figure*}[t]
    \centering
\includegraphics[width=\linewidth]{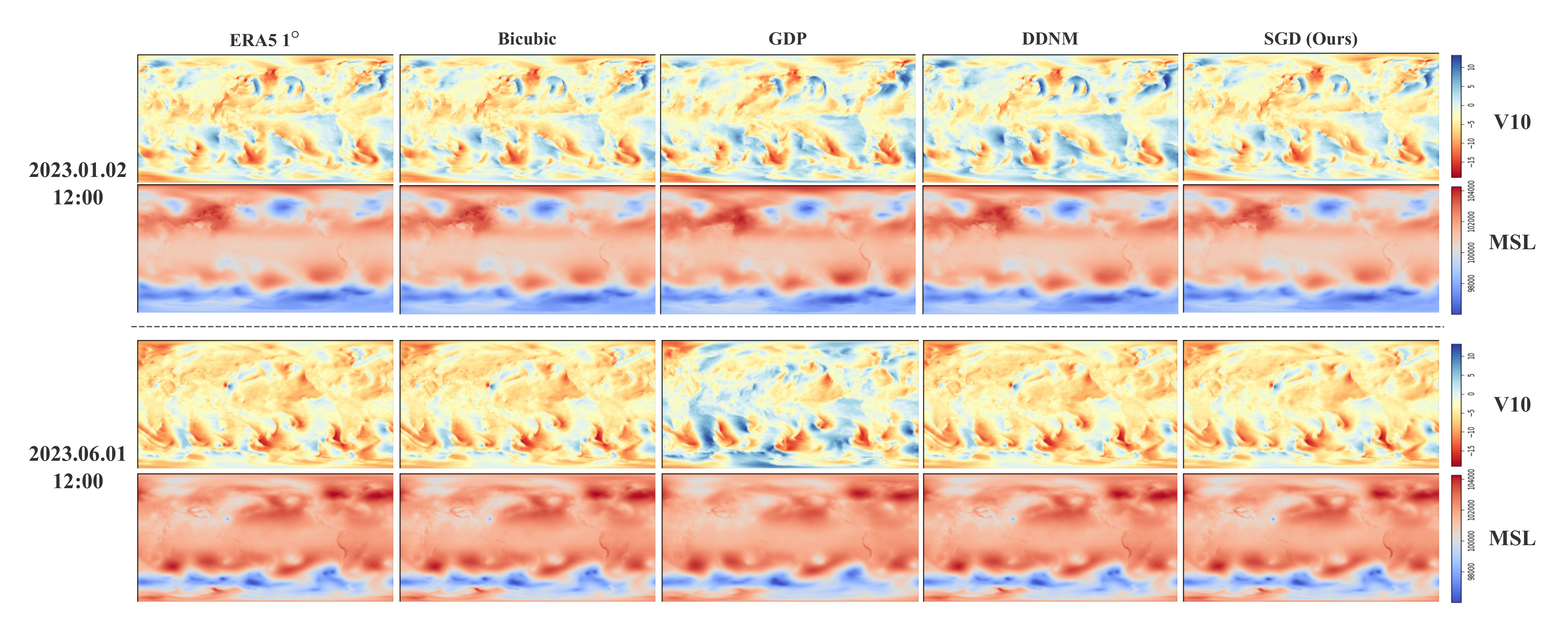}
    \caption{Visualization comparison of different interpolation-based and diffusion-based downscaling results in various time stamps. 
    We use different colors to distinguish $V_{10}$ and $MSL$. }
    \label{fig:app_results}
\end{figure*}

In this section, we present the downscaling results of SGD for the variables $V_{10}$ and $MSL$. 
\cref{fig:app_results} shows that SGD exhibits more faithful details in the maps as compared to interpolation-based and diffusion-based methods. 
Furthermore, SGD exhibits no discernible disparity in overall intensity when compared to ERA5 at a scale of $25km\times 25km$. 
Combining the results of the other two variables in the main text, it is validated that SGD is capable of producing highly satisfactory downscaling results across all four variables. 

%% file: app/ablation.tex
\section{Station observation-guided downscaling bias with stations in Weather2K}
In this section, we endeavor to integrate the MSE loss from ERA5 LR maps and MAE loss from the observation stations in Weather5K within the distance function utilized in the sampling process. 
Subsequently, we evaluated the high-resolution ERA5 maps derived from SGD with this setting across all stations within the Weather2K dataset, thereby further assessing the efficacy of the guided sampling and the accuracy of the downscaling results. 

Weather2K dataset~\cite{zhu2023weather2k} is a benchmark dataset that aims to address the shortcomings of existing weather forecasting datasets in terms of real-time relevance, reliability, and diversity, as well as the critical impediment posed by data quality. 
The data is available from January 2017 to August 2021. 
It encompasses the meteorological data from 2130 ground weather stations across 40896 time steps, with each data incorporates 3 position variables and 20 meteorological variables. 

Specifically, we incorporate the MAE loss between the generated HR ERA5 maps and station observations from the Weather5K dataset~\cite{han2024weather} with equal weights into our distance function to measure bias. 
Subsequently, we calculate the biases between the downscaling results obtained under this setting with the meteorological data at the stations from the Weather2K dataset. 
The evaluation metrics we employed are the MSE and MAE loss of the variable $T_{2M}$. 

We compared our results with those of interpolation-based and diffusion-based methods using the same metrics. 
As shown in ~\cref{tab:app_ablation}, the discrepancy between ERA5+station guided SGD and Weather2K stations is smaller, indicating that using ERA5 and Weather5K with equal weights as the distance function yields more ideal downscaling results for stations beyond Weather5K. 

\begin{figure*}[t]
    \centering
\includegraphics[width=\linewidth]{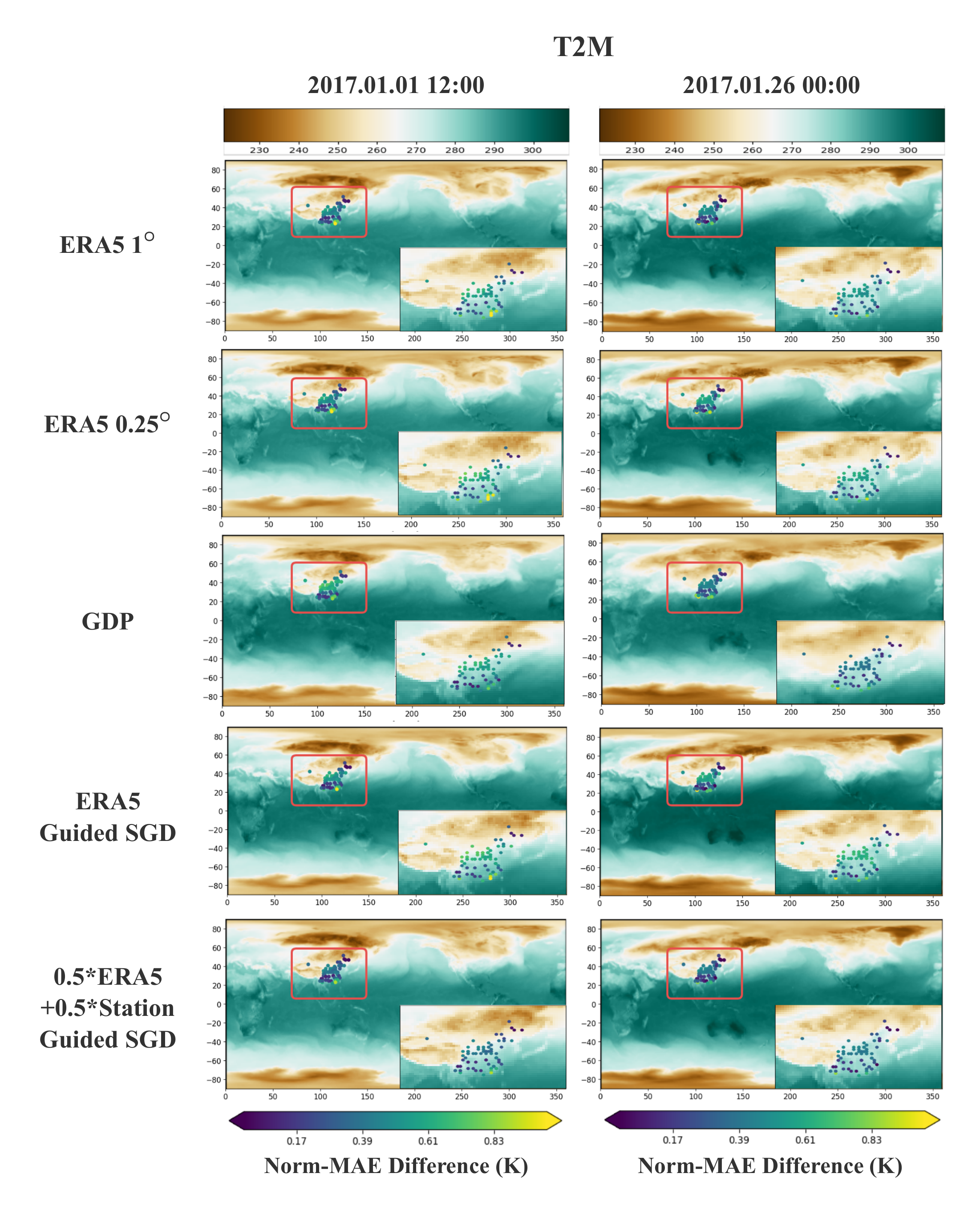}
    \caption{Visualization comparison of SGD downscaling to station-scale employing various distance function, where the coloration of each Weather2k observation station signifies the MAE loss between the downscaled results and their corresponding observed values. }
    \label{fig:app_ablation}
\end{figure*}

~\cref{fig:app_ablation} illustrates the differences in downscaling results among various methods at parts of the stations within Weather2K, with darker colors indicating smaller discrepancies at the stations. 
In terms of the bias between the downscaled results at the station locations in the image and the actual observations,  SGD with mixed guidance downscaling results has less extreme bias stations, which is symbolized as yellow-labeled stations. 
Moreover, the overall station coloration appears deeper. 
This suggests that utilizing weather5k as guidance can enhance the model's performance in downscaling at the local scale, aligning more closely with the real conditions.

\begin{table*}[t]
\centering
\caption{Station-level downscaling results for $T_{2M}$, which utilize the stations from Weather2k to assess the bias between the downscaling maps and Weather2k station observation values. 
ERA5 guided and ERA5 + station guided SGD respectively denote the SGD models that employ the MSE loss between the generated maps and ERA5 maps as the sole distance function, and the SGD model that integrates the Weather5k station observations into its distance function. }
\begin{tabular}{c| c| c c c c c}
    \toprule[1pt]
     Variable&Metrics&ERA5 1$^\circ$&ERA5 0.25$^\circ$&GDP &ERA5 Guided SGD&ERA5+Station Guided SGD\\
    \midrule
    \multirow{2}{*}{$T_{2M}$}
    &MSE& 17.51 & 17.80 & 18.87 & 18.08 & \textbf{15.61}\\
    &MAE& 407.81 & 420.38 & 466.00 & 431.33 & \textbf{355.31}\\
    \bottomrule[1pt]
  \end{tabular}

 \label{tab:app_ablation}
\end{table*}

%% file: app/running_time.tex
\section{Running Time and Resource Consumption}
\cref{time} shows the running time and resource consumption of SGD during the training and the sampling process. 
To enhance the inference efficiency, we have also tested our SGD on 50-step DDIM sampling to generate HR ERA5 maps within one minute, making it a feasible approach for practical use.
The efficiency and performance of DDIM sampling will be added to the revision.

\begin{table}[t]
\centering
\caption{The running time and resource consumption of SGD.}
\vspace{-0.2cm}
\resizebox{\linewidth}{!}{
\begin{tabular}{c|c c c}
    \toprule[1pt]
     Mode&SGD Training&SGD Sampling&SGD Sampling with DDIM\\
    \midrule
    \multirow{1}{*}{Running Time}
    &48h&6min&1min\\
    \midrule
    \multirow{1}{*}{Resource Consumption}
    &$5\times10^4$MiB&$1.8\times10^4$MiB&$1.6\times10^4$MiB\\
    \bottomrule[1pt]
  \end{tabular}
  }
 \label{time}
 \vspace{-0.2cm}
\end{table}

%% file: app/ablation_variables.tex
\section{Ablation Studies on the Relationship of Variables}
When only the brightness temperature variables from GridSat (IrWin\_Cdr or IrWin\_VZA\_Adj) are employed as the condition, a satisfactory performance can be obtained, demonstrating that brightness temperature is an important condition for ERA5 maps downscaling (\cref{table:gridsat}).
All variables from GridSat could guide the SGD to yield higher-quality HR ERA5 maps.
When incorporating GridSat as conditions and utilizing only a single LR ERA5 variable as guidance to generate the single HR ERA5 variable, the introduction of GridSat yields the most significant enhancement for the ERA5 temperature variable ($T_{2m}$) (\cref{table:era5}).
Considering the correlation between sea-level pressure and brightness temperature, the incorporation of GridSat as conditions for generating the single $MSL$ variable also contributes to enhancing the $MSL$ downscaling.

\begin{table}[t]\small
  \centering
  \caption{Ablation study employing a single GridSat variable.}
  \vspace{-0.2cm}
  \label{table:gridsat}
  \resizebox{\linewidth}{!}{
  \begin{tabular}{c|c c |c c |c c |c c }
    \toprule[1pt]
     \multirow{2}{*}{Methods}&\multicolumn{2}{c|}{$U_{10}$}&\multicolumn{2}{c|}{$V_{10}$}&\multicolumn{2}{c|}{$T_{2m}$}&\multicolumn{2}{c}{$MSL$}\\
     \cmidrule(lr){2-9}
     
    &MSE&MAE&MSE&MAE&MSE&MAE&MSE&MAE\\
    \midrule
Only IrWin\_Cdr  & 55.74 & 6.07 & 46.11 & 5.76 & 191.42 & 11.04 & 412.11 & 16.70 \\
Only IrWin\_VZA\_Adj  & 56.08 & 5.99 & 47.53 & 5.84 & 194.05 & 10.87 & 405.74 & 16.55 \\
Only IrWVP  & 60.42 & 6.74 & 55.08 & 6.10 & 214.07 & 12.07 & 424.17 & 17.52 \\
\midrule
All Variables in GridSat& \textbf{51.65} & \textbf{5.84} & \textbf{39.82} & \textbf{5.05} & \textbf{187.69} & \textbf{10.63} & \textbf{374.39} & \textbf{14.49} \\
    \bottomrule[1pt]

  \end{tabular}
  }
\vspace{-0.2cm}
\end{table}

\begin{table}[t]\small
\centering
\caption{Ablation study employing a single ERA5 variable.}
\vspace{-0.2cm}
\resizebox{\linewidth}{!}{
\begin{tabular}{c|c c | c | c c |c|c c | c | c c}
    \toprule[1pt]
     \multirow{2}{*}{Methods
     } &\multicolumn{2}{c|}{$U_{10}$} & \multirow{2}{*}{Methods}&\multicolumn{2}{c|}{$V_{10}$}&\multirow{2}{*}{Methods
     } &\multicolumn{2}{c|}{$T_{2m}$} & \multirow{2}{*}{Methods}&\multicolumn{2}{c}{$MSL$}\\
     \cmidrule(lr){2-3}
     \cmidrule(lr){5-6}
     \cmidrule(lr){8-9}
     \cmidrule(lr){11-12}

    & MSE & MAE    &&   MSE & MAE && MSE & MAE    &&   MSE & MAE  \\
    \midrule
    ERA5 $1^{\circ} $  & 53.18 & 5.95 & Era5 $1^{\circ} $ & 38.51 & 4.95 &  ERA5 $1^{\circ} $   & 216.27 & 11.39 & Era5 $1^{\circ} $ & 470.06 & 15.78 \\
    Only $U_{10}$   & 56.72 & 6.45 & Only $V_{10}$  & 47.28 & 5.84 & Only $T_{2m}$   & 194.15 & 10.71 & Only $MSL$  & 398.05 & 14.90   \\

    \bottomrule[1pt]

  \end{tabular}
  }
\label{table:era5}
\vspace{-0.3cm}
\end{table}

%% file: main.bbl
\begin{thebibliography}{47}
\providecommand{\natexlab}[1]{#1}
\providecommand{\url}[1]{\texttt{#1}}
\expandafter\ifx\csname urlstyle\endcsname\relax
  \providecommand{\doi}[1]{doi: #1}\else
  \providecommand{\doi}{doi: \begingroup \urlstyle{rm}\Url}\fi

\bibitem[Aich et~al.(2024)Aich, Hess, Pan, Bathiany, Huang, and Boers]{aich2024conditional}
Michael Aich, Philipp Hess, Baoxiang Pan, Sebastian Bathiany, Yu Huang, and Niklas Boers.
\newblock Conditional diffusion models for downscaling \& bias correction of earth system model precipitation.
\newblock \emph{arXiv preprint arXiv:2404.14416}, 2024.

\bibitem[Bischoff and Deck(2024)]{bischoff2024unpaired}
Tobias Bischoff and Katherine Deck.
\newblock Unpaired downscaling of fluid flows with diffusion bridges.
\newblock \emph{Artificial Intelligence for the Earth Systems}, 3\penalty0 (2):\penalty0 e230039, 2024.

\bibitem[Chajaei and Bagheri(2024)]{chajaei2024machine}
Fatemeh Chajaei and Hossein Bagheri.
\newblock Machine learning framework for high-resolution air temperature downscaling using lidar-derived urban morphological features.
\newblock \emph{Urban Climate}, 57:\penalty0 102102, 2024.

\bibitem[Chaudhuri and Robertson(2020)]{chaudhuri2020cligan}
Chiranjib Chaudhuri and Colin Robertson.
\newblock Cligan: A structurally sensitive convolutional neural network model for statistical downscaling of precipitation from multi-model ensembles.
\newblock \emph{Water}, 12\penalty0 (12):\penalty0 3353, 2020.

\bibitem[Chen et~al.(2023{\natexlab{a}})Chen, Du, Hu, Wang, and Wang]{chen2023swinrdm}
Lei Chen, Fei Du, Yuan Hu, Zhibin Wang, and Fan Wang.
\newblock Swinrdm: integrate swinrnn with diffusion model towards high-resolution and high-quality weather forecasting.
\newblock In \emph{Proceedings of the AAAI Conference on Artificial Intelligence}, pages 322--330, 2023{\natexlab{a}}.

\bibitem[Chen et~al.(2023{\natexlab{b}})Chen, Long, Jiang, Liu, and Zhang]{chen2023foundation}
Shengchao Chen, Guodong Long, Jing Jiang, Dikai Liu, and Chengqi Zhang.
\newblock Foundation models for weather and climate data understanding: A comprehensive survey.
\newblock \emph{arXiv preprint arXiv:2312.03014}, 2023{\natexlab{b}}.

\bibitem[Chen et~al.(2020)Chen, Zhao, Jia, Cao, and Liu]{chen2020adversarial}
Yuan Chen, Yang Zhao, Wei Jia, Li Cao, and Xiaoping Liu.
\newblock Adversarial-learning-based image-to-image transformation: A survey.
\newblock \emph{Neurocomputing}, 411:\penalty0 468--486, 2020.

\bibitem[De~Caceres et~al.(2018)De~Caceres, Martin-StPaul, Turco, Cabon, and Granda]{de2018estimating}
Miquel De~Caceres, Nicolas Martin-StPaul, Marco Turco, Antoine Cabon, and Victor Granda.
\newblock Estimating daily meteorological data and downscaling climate models over landscapes.
\newblock \emph{Environmental Modelling \& Software}, 108:\penalty0 186--196, 2018.

\bibitem[Dhariwal and Nichol(2021)]{dhariwal2021diffusion}
Prafulla Dhariwal and Alexander Nichol.
\newblock Diffusion models beat gans on image synthesis.
\newblock \emph{Advances in Neural Information Processing Systems}, 34:\penalty0 8780--8794, 2021.

\bibitem[D{\"o}ll et~al.(2024)D{\"o}ll, Abbasi, Messager, Trautmann, Lehner, and Lamouroux]{doll2024streamflow}
Petra D{\"o}ll, Mahdi Abbasi, Mathis~Lo{\"\i}c Messager, Tim Trautmann, Bernhard Lehner, and Nicolas Lamouroux.
\newblock Streamflow intermittence in europe: Estimating high-resolution monthly time series by downscaling of simulated runoff and random forest modeling.
\newblock \emph{Water Resources Research}, 60\penalty0 (8):\penalty0 e2023WR036900, 2024.

\bibitem[Fang et~al.(2013)Fang, Du, Xu, Shi, Li, and Ming]{fang2013spatial}
Jian Fang, Juan Du, Wei Xu, Peijun Shi, Man Li, and Xiaodong Ming.
\newblock Spatial downscaling of trmm precipitation data based on the orographical effect and meteorological conditions in a mountainous area.
\newblock \emph{Advances in Water Resources}, 61:\penalty0 42--50, 2013.

\bibitem[Fei et~al.(2023)Fei, Lyu, Pan, Zhang, Yang, Luo, Zhang, and Dai]{fei2023generative}
Ben Fei, Zhaoyang Lyu, Liang Pan, Junzhe Zhang, Weidong Yang, Tianyue Luo, Bo Zhang, and Bo Dai.
\newblock Generative diffusion prior for unified image restoration and enhancement.
\newblock In \emph{Proceedings of the IEEE/CVF Conference on Computer Vision and Pattern Recognition}, pages 9935--9946, 2023.

\bibitem[Feng et~al.(2022)Feng, He, Fu, Jampani, Akula, Narayana, Basu, Wang, and Wang]{feng2022training}
Weixi Feng, Xuehai He, Tsu-Jui Fu, Varun Jampani, Arjun Akula, Pradyumna Narayana, Sugato Basu, Xin~Eric Wang, and William~Yang Wang.
\newblock Training-free structured diffusion guidance for compositional text-to-image synthesis.
\newblock \emph{arXiv preprint arXiv:2212.05032}, 2022.

\bibitem[Gao et~al.(2023)Gao, Liu, Zeng, Xu, Li, Luo, Liu, Zhen, and Zhang]{gao2023implicit}
Sicheng Gao, Xuhui Liu, Bohan Zeng, Sheng Xu, Yanjing Li, Xiaoyan Luo, Jianzhuang Liu, Xiantong Zhen, and Baochang Zhang.
\newblock Implicit diffusion models for continuous super-resolution.
\newblock In \emph{Proceedings of the IEEE/CVF conference on computer vision and pattern recognition}, pages 10021--10030, 2023.

\bibitem[Han et~al.(2024)Han, Guo, Chen, Xu, and Bai]{han2024weather}
Tao Han, Song Guo, Zhenghao Chen, Wanghan Xu, and Lei Bai.
\newblock Weather-5k: A large-scale global station weather dataset towards comprehensive time-series forecasting benchmark.
\newblock \emph{arXiv preprint arXiv:2406.14399}, 2024.

\bibitem[Hersbach et~al.(2020)Hersbach, Bell, Berrisford, Hirahara, Hor{\'a}nyi, Mu{\~n}oz-Sabater, Nicolas, Peubey, Radu, Schepers, et~al.]{hersbach2020era5}
Hans Hersbach, Bill Bell, Paul Berrisford, Shoji Hirahara, Andr{\'a}s Hor{\'a}nyi, Joaqu{\'\i}n Mu{\~n}oz-Sabater, Julien Nicolas, Carole Peubey, Raluca Radu, Dinand Schepers, et~al.
\newblock The era5 global reanalysis.
\newblock \emph{Quarterly Journal of the Royal Meteorological Society}, 146\penalty0 (730):\penalty0 1999--2049, 2020.

\bibitem[Hess et~al.(2024)Hess, Aich, Pan, and Boers]{hess2024fast}
Philipp Hess, Michael Aich, Baoxiang Pan, and Niklas Boers.
\newblock Fast, scale-adaptive, and uncertainty-aware downscaling of earth system model fields with generative foundation models.
\newblock \emph{arXiv preprint arXiv:2403.02774}, 2024.

\bibitem[Ho et~al.(2020)Ho, Jain, and Abbeel]{ho2020denoising}
Jonathan Ho, Ajay Jain, and Pieter Abbeel.
\newblock Denoising diffusion probabilistic models.
\newblock \emph{Advances in neural information processing systems}, 33:\penalty0 6840--6851, 2020.

\bibitem[Ho et~al.(2022)Ho, Saharia, Chan, Fleet, Norouzi, and Salimans]{ho2022cascaded}
Jonathan Ho, Chitwan Saharia, William Chan, David~J Fleet, Mohammad Norouzi, and Tim Salimans.
\newblock Cascaded diffusion models for high fidelity image generation.
\newblock \emph{Journal of Machine Learning Research}, 23\penalty0 (47):\penalty0 1--33, 2022.

\bibitem[Huang et~al.(2022)Huang, Zhang, Fu, Chen, Zhang, Wang, and He]{huang2022task}
Yixuan Huang, Xiaoyun Zhang, Yu Fu, Siheng Chen, Ya Zhang, Yan-Feng Wang, and Dazhi He.
\newblock Task decoupled framework for reference-based super-resolution.
\newblock In \emph{Proceedings of the IEEE/CVF Conference on Computer Vision and Pattern Recognition}, pages 5931--5940, 2022.

\bibitem[Huang et~al.(2019)Huang, Wang, Huang, Huang, Wei, and Liu]{huang2019ccnet}
Zilong Huang, Xinggang Wang, Lichao Huang, Chang Huang, Yunchao Wei, and Wenyu Liu.
\newblock Ccnet: Criss-cross attention for semantic segmentation.
\newblock In \emph{Proceedings of the IEEE/CVF international conference on computer vision}, pages 603--612, 2019.

\bibitem[Jiang et~al.(2021)Jiang, Chan, Wang, Loy, and Liu]{jiang2021robust}
Yuming Jiang, Kelvin~CK Chan, Xintao Wang, Chen~Change Loy, and Ziwei Liu.
\newblock Robust reference-based super-resolution via c2-matching.
\newblock In \emph{Proceedings of the IEEE/CVF Conference on Computer Vision and Pattern Recognition}, pages 2103--2112, 2021.

\bibitem[Li et~al.(2022)Li, Yang, Chang, Chen, Feng, Xu, Li, and Chen]{li2022srdiff}
Haoying Li, Yifan Yang, Meng Chang, Shiqi Chen, Huajun Feng, Zhihai Xu, Qi Li, and Yueting Chen.
\newblock Srdiff: Single image super-resolution with diffusion probabilistic models.
\newblock \emph{Neurocomputing}, 479:\penalty0 47--59, 2022.

\bibitem[Li et~al.(2024)Li, Liu, Chen, Chen, Liang, Zou, and Shi]{li2024deepphysinet}
Wenyuan Li, Zili Liu, Keyan Chen, Hao Chen, Shunlin Liang, Zhengxia Zou, and Zhenwei Shi.
\newblock Deepphysinet: Bridging deep learning and atmospheric physics for accurate and continuous weather modeling.
\newblock \emph{arXiv preprint arXiv:2401.04125}, 2024.

\bibitem[Liu et~al.(2022)Liu, Yuan, Pan, Fu, Liu, and Lu]{liu2022diffusion}
Jinzhe Liu, Zhiqiang Yuan, Zhaoying Pan, Yiqun Fu, Li Liu, and Bin Lu.
\newblock Diffusion model with detail complement for super-resolution of remote sensing.
\newblock \emph{Remote Sensing}, 14\penalty0 (19):\penalty0 4834, 2022.

\bibitem[Liu et~al.(2024{\natexlab{a}})Liu, Dong, Pan, Dong, Chen, Zhang, Mei, Fu, and Wang]{liu2024patchscaler}
Yong Liu, Hang Dong, Jinshan Pan, Qingji Dong, Kai Chen, Rongxiang Zhang, Xing Mei, Lean Fu, and Fei Wang.
\newblock Patchscaler: An efficient patch-independent diffusion model for super-resolution.
\newblock \emph{arXiv preprint arXiv:2405.17158}, 2024{\natexlab{a}}.

\bibitem[Liu et~al.(2024{\natexlab{b}})Liu, Chen, Bai, Li, Chen, Wang, Ouyang, Zou, and Shi]{liu2024deriving}
Zili Liu, Hao Chen, Lei Bai, Wenyuan Li, Keyan Chen, Zhengyi Wang, Wanli Ouyang, Zhengxia Zou, and Zhenwei Shi.
\newblock Deriving accurate surface meteorological states at arbitrary locations via observation-guided continous neural field modeling.
\newblock \emph{IEEE Transactions on Geoscience and Remote Sensing}, 2024{\natexlab{b}}.

\bibitem[Liu et~al.(2024{\natexlab{c}})Liu, Chen, Bai, Li, Chen, Wang, Ouyang, Zou, and Shi]{liu2024observation}
Zili Liu, Hao Chen, Lei Bai, Wenyuan Li, Keyan Chen, Zhengyi Wang, Wanli Ouyang, Zhengxia Zou, and Zhenwei Shi.
\newblock Observation-guided meteorological field downscaling at station scale: A benchmark and a new method.
\newblock \emph{arXiv preprint arXiv:2401.11960}, 2024{\natexlab{c}}.

\bibitem[Liu et~al.(2024{\natexlab{d}})Liu, Chen, Bai, Li, Ouyang, Zou, and Shi]{liu2024mambads}
Zili Liu, Hao Chen, Lei Bai, Wenyuan Li, Wanli Ouyang, Zhengxia Zou, and Zhenwei Shi.
\newblock Mambads: Near-surface meteorological field downscaling with topography constrained selective state space modeling.
\newblock \emph{arXiv preprint arXiv:2408.10854}, 2024{\natexlab{d}}.

\bibitem[Lopez-Gomez et~al.(2024)Lopez-Gomez, Wan, Zepeda-N{\'u}{\~n}ez, Schneider, Anderson, and Sha]{lopez2024dynamical}
Ignacio Lopez-Gomez, Zhong~Yi Wan, Leonardo Zepeda-N{\'u}{\~n}ez, Tapio Schneider, John Anderson, and Fei Sha.
\newblock Dynamical-generative downscaling of climate model ensembles.
\newblock \emph{arXiv preprint arXiv:2410.01776}, 2024.

\bibitem[Maraun(2016)]{maraun2016bias}
Douglas Maraun.
\newblock Bias correcting climate change simulations-a critical review.
\newblock \emph{Current Climate Change Reports}, 2\penalty0 (4):\penalty0 211--220, 2016.

\bibitem[McNally et~al.(2024)McNally, Lessig, Lean, Boucher, Alexe, Pinnington, Chantry, Lang, Burrows, Chrust, et~al.]{mcnally2024data}
Anthony McNally, Christian Lessig, Peter Lean, Eulalie Boucher, Mihai Alexe, Ewan Pinnington, Matthew Chantry, Simon Lang, Chris Burrows, Marcin Chrust, et~al.
\newblock Data driven weather forecasts trained and initialised directly from observations.
\newblock \emph{arXiv preprint arXiv:2407.15586}, 2024.

\bibitem[Menon et~al.(2020)Menon, Damian, Hu, Ravi, and Rudin]{menon2020pulse}
Sachit Menon, Alexandru Damian, Shijia Hu, Nikhil Ravi, and Cynthia Rudin.
\newblock Pulse: Self-supervised photo upsampling via latent space exploration of generative models.
\newblock In \emph{Proceedings of the ieee/cvf conference on computer vision and pattern recognition}, pages 2437--2445, 2020.

\bibitem[Mukkavilli et~al.(2023)Mukkavilli, Civitarese, Schmude, Jakubik, Jones, Nguyen, Phillips, Roy, Singh, Watson, et~al.]{mukkavilli2023ai}
S~Karthik Mukkavilli, Daniel~Salles Civitarese, Johannes Schmude, Johannes Jakubik, Anne Jones, Nam Nguyen, Christopher Phillips, Sujit Roy, Shraddha Singh, Campbell Watson, et~al.
\newblock Ai foundation models for weather and climate: Applications, design, and implementation.
\newblock \emph{arXiv preprint arXiv:2309.10808}, 2023.

\bibitem[Pan et~al.(2021)Pan, Zhan, Dai, Lin, Loy, and Luo]{pan2021exploiting}
Xingang Pan, Xiaohang Zhan, Bo Dai, Dahua Lin, Chen~Change Loy, and Ping Luo.
\newblock Exploiting deep generative prior for versatile image restoration and manipulation.
\newblock \emph{IEEE Transactions on Pattern Analysis and Machine Intelligence}, 44\penalty0 (11):\penalty0 7474--7489, 2021.

\bibitem[Pozo~Buil et~al.(2021)Pozo~Buil, Jacox, Fiechter, Alexander, Bograd, Curchitser, Edwards, Rykaczewski, and Stock]{pozo2021dynamically}
Mercedes Pozo~Buil, Michael~G Jacox, Jerome Fiechter, Michael~A Alexander, Steven~J Bograd, Enrique~N Curchitser, Christopher~A Edwards, Ryan~R Rykaczewski, and Charles~A Stock.
\newblock A dynamically downscaled ensemble of future projections for the california current system.
\newblock \emph{Frontiers in Marine Science}, 8:\penalty0 612874, 2021.

\bibitem[Qi et~al.(2023)Qi, Cun, Zhang, Lei, Wang, Shan, and Chen]{qi2023fatezero}
Chenyang Qi, Xiaodong Cun, Yong Zhang, Chenyang Lei, Xintao Wang, Ying Shan, and Qifeng Chen.
\newblock Fatezero: Fusing attentions for zero-shot text-based video editing.
\newblock In \emph{Proceedings of the IEEE/CVF International Conference on Computer Vision}, pages 15932--15942, 2023.

\bibitem[Skofronick-Jackson et~al.(2015)Skofronick-Jackson, Hudak, Petersen, Nesbitt, Chandrasekar, Durden, Gleicher, Huang, Joe, Kollias, et~al.]{skofronick2015global}
Gail Skofronick-Jackson, David Hudak, Walter Petersen, Stephen~W Nesbitt, V Chandrasekar, Stephen Durden, Kirstin~J Gleicher, Gwo-Jong Huang, Paul Joe, Pavlos Kollias, et~al.
\newblock Global precipitation measurement cold season precipitation experiment (gcpex): For measurement’s sake, let it snow.
\newblock \emph{Bulletin of the American Meteorological Society}, 96\penalty0 (10):\penalty0 1719--1741, 2015.

\bibitem[Sun et~al.(2024)Sun, Deng, Ren, Liu, Deng, and Jin]{sun2024deep}
Yongjian Sun, Kefeng Deng, Kaijun Ren, Jia Liu, Chongjiu Deng, and Yongjun Jin.
\newblock Deep learning in statistical downscaling for deriving high spatial resolution gridded meteorological data: A systematic review.
\newblock \emph{ISPRS Journal of Photogrammetry and Remote Sensing}, 208:\penalty0 14--38, 2024.

\bibitem[Tu et~al.(2024)Tu, Yang, and Fei]{tu2024taming}
Siwei Tu, Weidong Yang, and Ben Fei.
\newblock Taming generative diffusion for universal blind image restoration.
\newblock \emph{arXiv preprint arXiv:2408.11287}, 2024.

\bibitem[Vandal et~al.(2024)Vandal, Duffy, McDuff, Nachmany, and Hartshorn]{vandal2024global}
Thomas~J Vandal, Kate Duffy, Daniel McDuff, Yoni Nachmany, and Chris Hartshorn.
\newblock Global atmospheric data assimilation with multi-modal masked autoencoders.
\newblock \emph{arXiv preprint arXiv:2407.11696}, 2024.

\bibitem[Vaughan et~al.(2024)Vaughan, Markou, Tebbutt, Requeima, Bruinsma, Andersson, Herzog, Lane, Chantry, Hosking, et~al.]{vaughan2024aardvark}
Anna Vaughan, Stratis Markou, Will Tebbutt, James Requeima, Wessel~P Bruinsma, Tom~R Andersson, Michael Herzog, Nicholas~D Lane, Matthew Chantry, J~Scott Hosking, et~al.
\newblock Aardvark weather: end-to-end data-driven weather forecasting.
\newblock \emph{arXiv preprint arXiv:2404.00411}, 2024.

\bibitem[Wang et~al.(2021)Wang, Mao, Yuan, Shi, Cao, Qin, Duan, and Tang]{wang2021method}
Han Wang, Kebiao Mao, Zijin Yuan, Jiancheng Shi, Mengmeng Cao, Zhihao Qin, Sibo Duan, and Bohui Tang.
\newblock A method for land surface temperature retrieval based on model-data-knowledge-driven and deep learning.
\newblock \emph{Remote sensing of environment}, 265:\penalty0 112665, 2021.

\bibitem[Wang et~al.(2022)Wang, Yu, and Zhang]{wang2022zero}
Yinhuai Wang, Jiwen Yu, and Jian Zhang.
\newblock Zero-shot image restoration using denoising diffusion null-space model.
\newblock \emph{arXiv preprint arXiv:2212.00490}, 2022.

\bibitem[Wang et~al.(2024)Wang, Li, Cui, Cui, Xu, Hora, Zaveri, Rodella, Bai, and Long]{wang2024spatial}
Yiming Wang, Chen Li, Yingjie Cui, Yanhong Cui, Yuancheng Xu, Tejasvi Hora, Esha Zaveri, Aude-Sophie Rodella, Liangliang Bai, and Di Long.
\newblock Spatial downscaling of grace-derived groundwater storage changes across diverse climates and human interventions with random forests.
\newblock \emph{Journal of Hydrology}, 640:\penalty0 131708, 2024.

\bibitem[Zhu et~al.(2023)Zhu, Xiong, Wu, Nie, Zhang, and Yang]{zhu2023weather2k}
Xun Zhu, Yutong Xiong, Ming Wu, Gaozhen Nie, Bin Zhang, and Ziheng Yang.
\newblock Weather2k: A multivariate spatio-temporal benchmark dataset for meteorological forecasting based on real-time observation data from ground weather stations.
\newblock \emph{arXiv preprint arXiv:2302.10493}, 2023.

\bibitem[Zhu et~al.(2024)Zhu, Bo, Sun, Zhang, Sun, Shen, Zhang, Tang, Cao, and Wang]{zhu2024downscaling}
Zhongzheng Zhu, Yanchen Bo, Tongtong Sun, Xiaoran Zhang, Mei Sun, Aojie Shen, Yusha Zhang, Jia Tang, Mengfan Cao, and Chenyu Wang.
\newblock A downscaling-and-fusion framework for generating spatio-temporally complete and fine resolution remotely sensed surface soil moisture.
\newblock \emph{Agricultural and Forest Meteorology}, 352:\penalty0 110044, 2024.

\end{thebibliography}
